\documentclass[journal,twoside,web]{ieeecolor}
\usepackage{tmi}
\usepackage{cite}
\usepackage{amsmath,amssymb,amsfonts}
\usepackage{algorithmic}
\usepackage{graphicx}
\usepackage{textcomp}
\usepackage{multirow}
\usepackage{pifont}
\usepackage{mathrsfs}
\usepackage{booktabs}
\usepackage{makecell}
\usepackage{diagbox}
\usepackage{threeparttable}

\usepackage[implicit=false]{hyperref}

\def\BibTeX{{\rm B\kern-.05em{\sc i\kern-.025em b}\kern-.08em
    T\kern-.1667em\lower.7ex\hbox{E}\kern-.125emX}}
\begin{document}
\title{SEINE: Structure Encoding and Interaction Network for Nuclei Instance Segmentation}
\author{Ye Zhang, Linghan Cai, Ziyue Wang, Yongbing Zhang, \IEEEmembership{Senior Member, IEEE}	
	%\IEEEmembership{Fellow, IEEE}, Second B. Author,and Third C. Author, Jr., \IEEEmembership{Member, IEEE}	
\thanks{This work was supported in part by the National Natural Science Foundation of China under 62031023\&62331011; in part by the Shenzhen Science and Technology Project under JCYJ20200109142808034\&GXWD20220818170353009, and in part by the Fundamental Research Funds for the Central Universities under  HIT.OCEF.2023050. \textit{(Corresponding author: Yongbing Zhang.)}}
\thanks{Ye Zhang, Linghan Cai, Ziyue Wang, and Yongbing Zhang are with the School of Computer Science and Technology, Harbin Institute of Technology, Shenzhen, 518055, China. (e-mails: zhangye94@stu.hit.edu.cn; cailh@stu.hit.edu.cn; 200111326@stu.hit.edu.cn; ybzhang08@hit.edu.cn).}}

\maketitle

\begin{abstract}
Nuclei instance segmentation in histopathological images is of great importance for biological analysis and cancer diagnosis but remains challenging for two reasons. (1) Similar visual presentation of intranuclear and extranuclear regions of chromophobe nuclei often causes under-segmentation, and (2) current methods lack the exploration of nuclei structure, resulting in fragmented instance predictions. To address these problems, this paper proposes a \textbf{S}tructure \textbf{E}ncoding and \textbf{I}nteraction \textbf{NE}twork, termed SEINE, which develops the structure modeling scheme of nuclei and exploits the structure similarity between nuclei to improve the integrality of each segmented instance. Concretely, SEINE introduces a contour-based structure encoding (SE) that considers the correlation between nuclei structure and semantics, realizing a reasonable representation of the nuclei structure. Based on the encoding, we propose a structure-guided attention (SGA) that takes the clear nuclei as prototypes to enhance the structure learning for the fuzzy nuclei. To strengthen the structural learning ability, a semantic feature fusion (SFF) is presented to boost the semantic consistency of semantic and structure branches. 
Furthermore, a position enhancement (PE) method is applied to suppress incorrect nuclei boundary predictions. Extensive experiments demonstrate the superiority of our approaches, and SEINE achieves state-of-the-art (SOTA) performance on four datasets. The code is available at \href{https://github.com/zhangye-zoe/SEINE}{https://github.com/zhangye-zoe/SEINE}.
\end{abstract}


\definecolor{highlight}{RGB}{47,138,217} 
\section{Introduction}
\label{sec:introduction}
Nuclear instance segmentation plays a crucial role in histopathological image analysis, which is widely regarded as the gold standard for cancer diagnosis, treatment, and prevention \cite{khened2021generalized, hollandi2022nucleus}. This is primarily because the shape, appearance, and distribution of nuclei offer valuable and interpretable features for pathology analysis. In modern healthcare, manual annotation of nuclei in pathological images is time-consuming and prone to individual variations. Fortunately, computational pathology has facilitated automated nuclei segmentation, improving the efficiency of medical workers \cite{xing2015automatic, kumar2017dataset, qu2020weakly}.

\begin{figure}[t]
	\centering
	\includegraphics[width=0.83\linewidth]{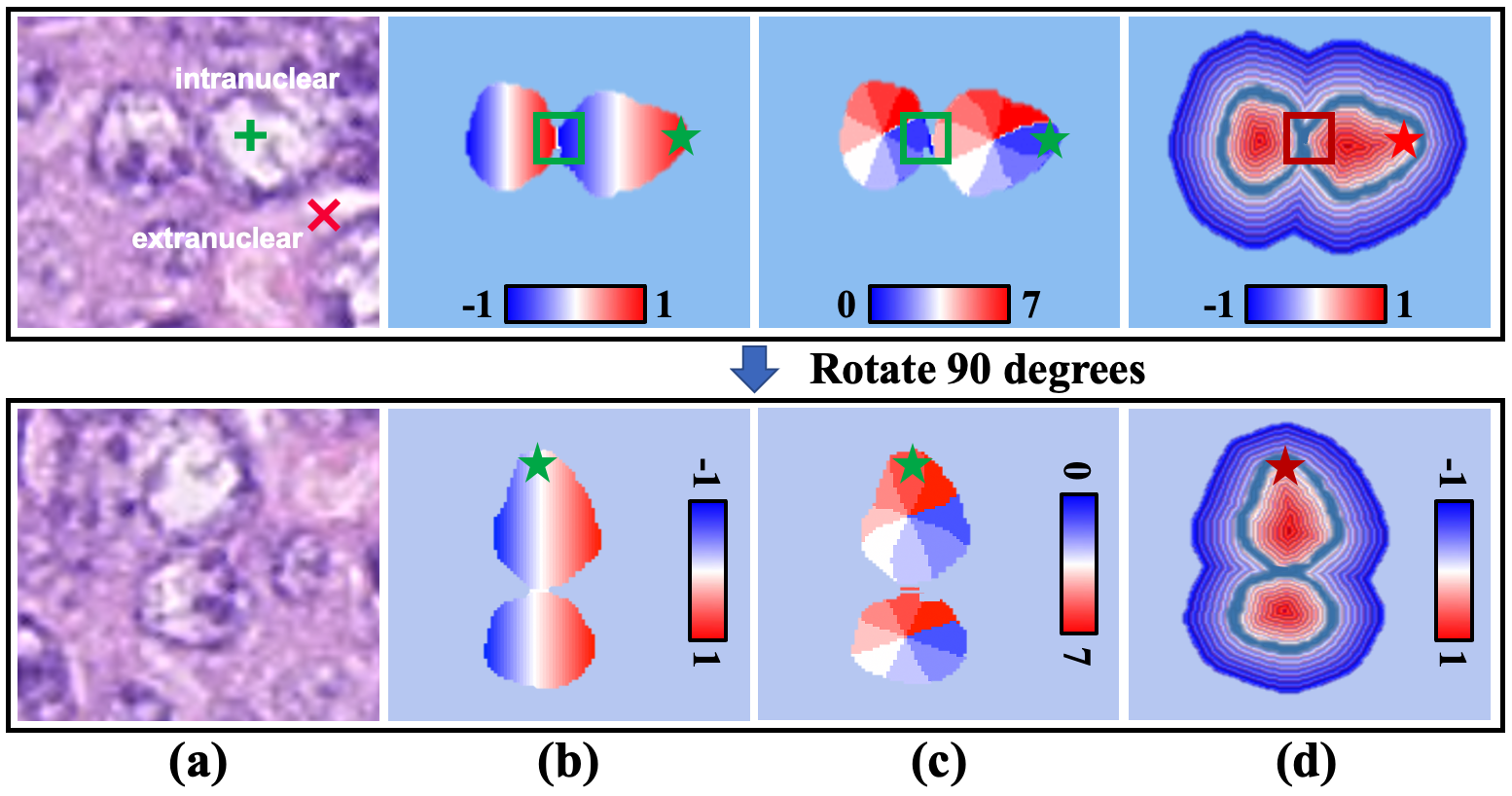}
	\caption{\textbf{Comparison of structure modeling methods}. (a) is the input image, (b) is the horizontal distance encoding, (c) is the centripetal direction encoding, and (d) is our proposed structure encoding (SE), whose contour is highlighted in the green line. The second row represents the structure modeling when using rotation data augmentation.
	}
	\label{fig:fig3}
	\vspace{-0.6cm}
\end{figure}

In recent years, deep learning techniques have prospered in the nuclei segmentation task \cite{liu2023learning,stringer2021cellpose, naylor2018segmentation }.
However, when faced with chromophobe nuclei, these methods suffer from under-segmentation for the following two reasons.
\textbf{Firstly}, the chromophobe nuclei generally exhibit high degrees of visual similarity to their cytoplasm and extranuclear matrix. This phenomenon arises from the close staining reaction in intranuclear (\textcolor{green}{\textbf{+}}) and extranuclear (\textcolor{red}{\textbf{×}}) regions of the chromophobe as shown in \textcolor{highlight}{\textbf{Fig. \ref{fig:fig3} (a)}}, posing difficulties in distinguishing the nuclei \cite{bettinger1991new, ross2006histology}.
\textbf{Secondly}, existing nuclei segmentation methods tend to utilize semantic features for pixel-by-pixel classification. However, the lack of attention to nuclei structure makes these methods powerless to differentiate the visually similar areas, resulting in incorrect semantic prediction. Nuclei structure interaction is a possible solution for the issue. In histopathological images, a tile usually has a number of nuclei that have similar size and shape. The clearly stained nuclei could promote the structure learning of chromophobe nuclei.

 A proper representation of nuclei structure is a prerequisite for achieving structure interaction. Currently, several methods have been proposed to measure nuclei structure. Centroid-based methods \cite{graham2019hover, he2021cdnet} encode nuclear regions based on their relative position to the centroid. These strategies assign discrepant encodings to adjacent regions belonging to different instances, thereby distinguishing the overlapping instances as shown in \textcolor{highlight}{\textbf{Fig. \ref{fig:fig3} (b)-(c)}}. Nevertheless, these approaches break the semantic consistency of similar structures. In detail, the pixels with the same distance to their centroids should keep the same structure, i.e., isotropic. As shown in green boxes, the horizontal distance encoding in \textcolor{highlight}{\textbf{Fig. \ref{fig:fig3} (b)}} set an opposite value to the similar structure and the centripetal direction encoding in \textcolor{highlight}{\textbf{Fig. \ref{fig:fig3} (c)}} assign the different categories to the similar regions. In addition, structural encoding, as an intrinsic feature of the nuclei, should not be affected by the orientated variations of input. However, as the green star indicates, when rotation data augmentation is applied, both types of structural encodings change accordingly. Hence, an isotropic and direction-invariant structure modeling method is necessary for measuring nuclei structures.

\begin{figure}
	\centering
	\includegraphics[width=0.82\linewidth]{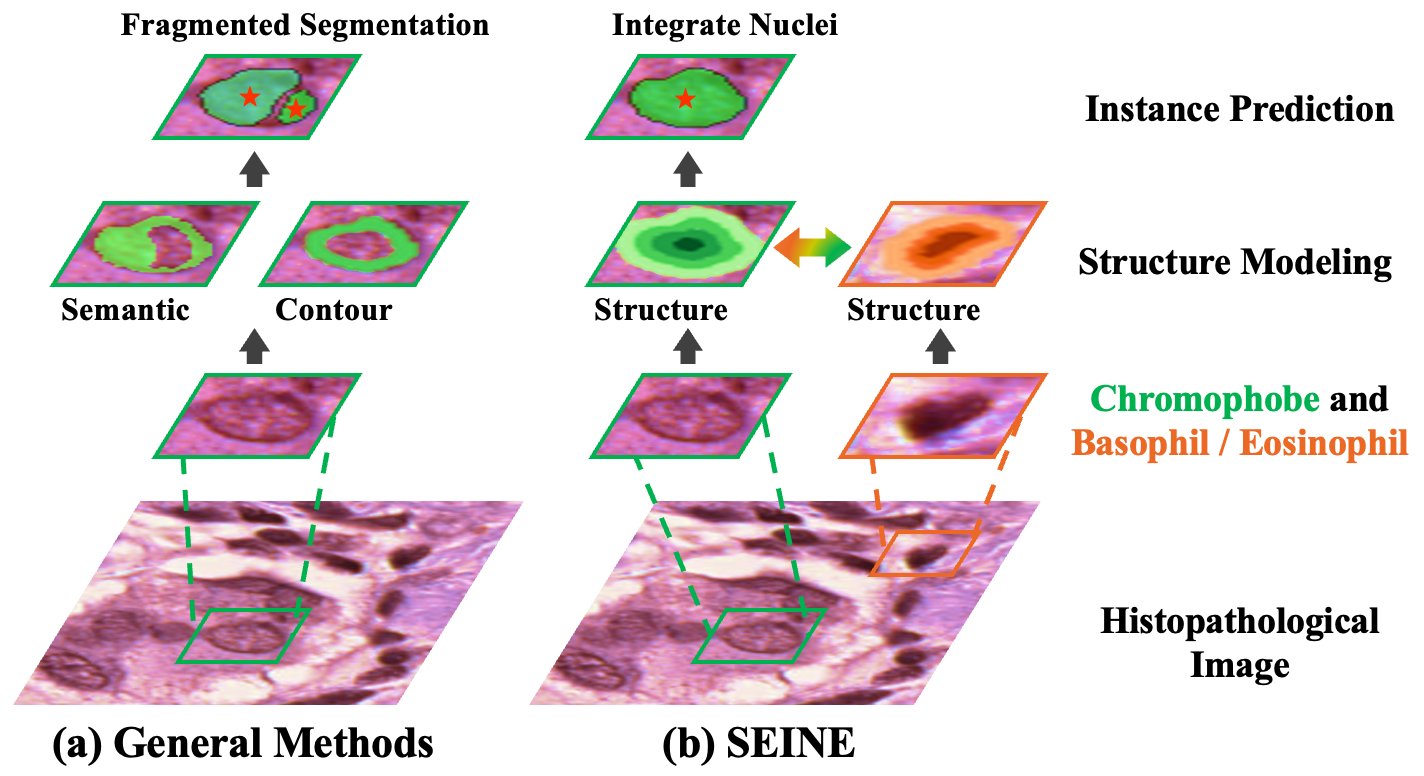}
	\caption{\textbf{The comparison of the general methods and our SEINE.} The third row exhibits a chromophobe and a clearly stained nucleus. In the chromophobe, the intranuclear region is ambiguous, but the membrane is distinct. In the clearly stained nucleus, both the intranuclear region and membrane are clear. The double-headed arrow in (b) represents the structure interaction.}
	\label{fig:fig1}
	\vspace{-0.5cm}
\end{figure}

At the same time, some methods \cite{zhou2019cia, chen2023cpp, han2023ensemble} make use of the clear membrane to model the nuclei structure. Despite great progress made in refining contour predictions, the identification of nuclear membrane alone is insufficient to distinguish the intranuclear and extranuclear regions, as it neglects structural relationships among the nucleus, the membrane, and the extranuclear tissues. This ignorance leads to holes appearing in the nuclei semantic prediction and makes a nucleus be mistakenly separated into multiple instances, as shown in \textcolor{highlight}{\textbf{Fig. \ref{fig:fig1} (a)}}. 
Therefore, it is imperative to develop a robust structural modeling strategy that effectively integrates the representation of nuclei at both the semantic and structural levels, guaranteeing consistency in semantic categories both within and outside the contour.

Considering the above problems, this paper proposes a contour-based structure encoding (SE) strategy as shown in \textcolor{highlight}{\textbf{Fig. \ref{fig:fig3} (d)}}. The SE is isotropic and direction-invariant, calculating the distances from all pixels to the contour. The degree of the distances describes the location of pixels to the entire nucleus, while the sign of distances indicates the semantic category. Notably, the SE not only encodes the nuclear region but incorporates the background, enabling a unified representation of semantics and structure.
To ensure the accurate structure learning of chromophobe nuclei and establish consistency between semantics and structure, we develop a cross-nuclei interaction scheme using a structure-guided attention (SGA) module. This approach leverages clearly stained nuclei as prototypes for the structure learning of chromophobe nuclei as depicted in \textcolor{highlight}{\textbf{Fig. \ref{fig:fig1} (b)}}. 
To further enhance the feature representation and suppress unreasonable contour prediction, we introduce two modules: semantic feature fusion (SFF) and position enhancement (PE). SFF concatenates the feature from the semantic branch to enhance the feature representation of the structure branch. The PE focuses on the centroid of nuclei and corrects centroid shift, promoting the integrality of nuclei.

In a nutshell, our proposed network is named \textbf{s}tructure \textbf{e}ncoding and \textbf{i}nteraction \textbf{ne}twork (SEINE) and our contributions can be summarised as following five points:

\begin{itemize}
	\item This paper highlights challenges of chromophobe nuclei segmentation for the first time and presents an effective solution termed SEINE, which focuses on the nuclei structures for solving the under-segmentation problem.
	
	\item This paper proposes an isotropic nuclei structural encoding strategy that bridges the gap between semantic and structural feature representations.
	
	\item A novel structure-guided attention mechanism is designed for enhancing structure interactions across nuclei, thereby improving the integrality of nuclei structure.
	
	\item To further enhance structure feature representation, we design a semantic feature fusion that incorporates semantic information for structure features. Additionally, a position enhancement module is developed that improves the network's perception of nuclei positions.
	
	\item Extensive experiments demonstrate the superiority of our method, and SEINE achieves state-of-the-art performances on four popular nuclei segmentation datasets. 
\end{itemize}

The remainders\ of this paper are organized as follows. The related work is illustrated in \textbf{\textcolor{highlight}{Section \ref{relate}}}. \textbf{\textcolor{highlight}{Section \ref{method}}} describes our SEINE in detail. Subsequently, extensive experiments and analyses are shown in \textbf{\textcolor{highlight}{Section \ref{exper}}}. The discussion and conclusion are presented in \textbf{\textcolor{highlight}{Section \ref{discuss}}} and \textbf{\textcolor{highlight}{Section \ref{conclu}}}, respectively.

\begin{figure*}
	\centering
	\includegraphics[width=0.85\linewidth]{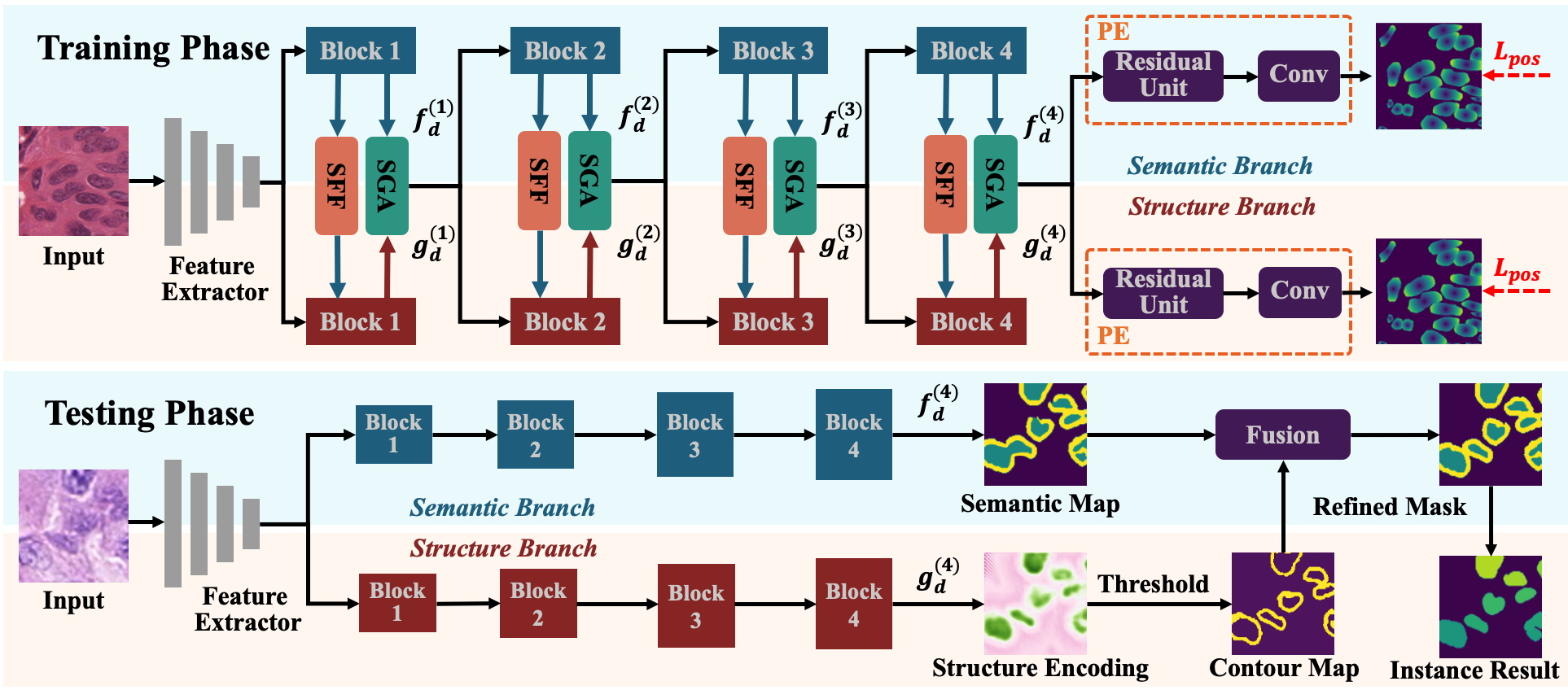}
	\caption{\textbf{The flowchart of the proposed SEINE.} The top and the bottom of the flowchart represent training phase and testing phase, repectively.  In the training phase, the network contains a shared feature extractor, a semantic branch, and a structure branch. For each block of the two branches, the semantic feature fusion (SFF) module and the structure-guided attention (SGA) module are constructed to perform structure interaction. The position enhancement (PE) is used to enhance the localization ability of nuclei centroid. In the testing phase, the prediction results from the semantic branch and structure branch are combined for refined instance results.}
	\label{fig:fig2}
	\vspace{-0.3cm}
\end{figure*}

\section{Related Work}\label{relate}
\subsection{Nuclei Instance Segmentation}
Traditional nuclei instance segmentation methods rely on threshold setting, such as watershed \cite{levner2007classification} and Otsu \cite{zhang2008image, feng2017multi}, which cannot adapt to the various scenes.
Since U-Net \cite{ronneberger2015u} won the cell tracking challenge, convolutional neural networks have been widely applied to nuclei instance segmentation. 
Currently, some studies take advantage of contour information to solve boundary-overlapping problems.
DCAN \cite{chen2016dcan} incorporates contour information into the segmentation branch. 
CIA-Net \cite{zhou2019cia} combines the spatial and textural features between nuclei and contours. 
%CRAC \cite{liu2023learning} segments biomedical images using the consistency constraint of instance boundaries on both explicit and implicit affinity maps. 
Meanwhile, several works measure distances based on the centroid to improve nuclei segmentation ability.  
HoverNet \cite{graham2019hover} obtains nuclei-centroid horizontal and vertical distances to enhance the model fitting ability. 
StarDist \cite{schmidt2018cell} distinguishes neighboring nuclei with $k$ radial directions from the nuclei centroid to the boundary. 
Additionally, DoNet \cite{jiang2023donet} employs a decompose-and-recombine strategy, which separately predicts the intersection and complements parts to enhance the segmentation capability.
However, these methods do not adequately characterize the structure, leading to an under-segmentation problem. 
By contrast, this paper investigates the nuclei structure modeling methods and proposes a contour-based structure encoding for obtaining a reasonable representation of nuclei structure.

\subsection{Attention in Medical Image Segmentation}
Due to the low contrast and blurry boundaries of medical images, numerous attention-based methods are proposed to enhance the segmentation performance.
%MedT \cite{valanarasu2021medical} introduces a gated axial-attention method to existing architectures, enabling the awareness of contextual features.
EANet \cite{wang2022eanet} employs a contour-based attention module, reducing feature noises and enhancing structure characteristics.
ResGANet \cite{cheng2022resganet} applies group attention blocks in both the channel and spatial dimensions to capture feature dependencies.
PistoSeg \cite{fang2023weakly} reduces the discrepancy between synthetic and real images by incorporating attention-based feature consistency constraints.
With the success of vision transformer \cite{dosovitskiy2020image}, researchers widely study the self-attention mechanism for medical image segmentation.
PPFormer \cite{cai2022using} utilizes a local-to-global self-attention mechanism that expands the receptive field and extracts valuable contextual information.
nnFormer \cite{zhou2023nnformer} introduces self-attention mechanisms to 3D medical image segmentation models for learning volumetric representations.
$\omega$-net \cite{xu2022omega} uses a multi-dimensional self-attention mechanism to highlight salient features and suppress irrelevant features.
In this paper, we propose a novel SGA that takes advantage of the powerful spatial modeling ability of self-attention for structure interaction between nuclei, effectively reducing the under-segmentation phenomenon in nuclei instance segmentation.

%In this paper, we propose a novel attention SGA that takes advantage of the powerful spatial modeling ability of self-attention for structure interaction between nuclei.

\section{Methodology}\label{method}
\subsection{Framework Overview}
The overall framework of SEINE is illustrated in  \textcolor{highlight}{\textbf{Fig. \ref{fig:fig2}}}, which consists of a four-stage feature extractor, a semantic branch, and a structure branch. The feature extractor extracts deep features. With these features, the semantic branch conducts a three-category segmentation task, distinguishing the foreground, background, and contour. Meanwhile, the structure branch carries out a contour-based distance prediction task to encode the nuclei structure. 
In the training phase (\textcolor{highlight}{\textbf{Fig. \ref{fig:fig2}}} top), we employ a structure-guided attention (SGA) module between the semantic branch and the structure branch to guide the structure interaction across nuclei. In the meanwhile, we devise a semantic feature fusion (SFF) module to help the structure branch aggregate features of the semantic branch, thereby obtaining better feature representations. To locate nuclei, position enhancement (PE) modules are adopted in the two branches to reduce incorrect contour predictions. 

The total loss function $L_{total}$ is generally defined as:
\begin{equation}
	L_{total} = L_{sem} + \lambda_1 L_{str} + \lambda_2 L_{pos}, 
\end{equation}
where $L_{sem}$ is the loss of the semantic branch, $L_{str}$ represents the loss of the structure branch, and $L_{pos}$ denotes the loss of the PE. These loss functions will be introduced in the following sub-section. In this paper, $\lambda_1$ and $\lambda_2$ are set to 1 for balancing the importance of each term.

In the testing phase (\textcolor{highlight}{\textbf{Fig. \ref{fig:fig2}}} bottom), we use the output from the structure branch to refine the predictions of the semantic branch. Specifically, the structure encoding prediction is filtered by a threshold for high-confidence contours, which are utilized to compensate for the contour-class semantic prediction, generating the final instance segmentation result.

\subsection{Contour-based Structure Encoding}

Existing structure modeling approaches typically model the structure by means of the centroid of nuclei, such as the \textbf{h}orizontal\&\textbf{v}ertical (HV) map  \cite{graham2019hover} or the \textbf{dir}ection (Dir) map \cite{he2021cdnet}.
However, these modeling strategies neglect the relationship between semantic categories and nuclei structures, resulting in semantic inconsistency between similar structures, as shown in \textcolor{highlight}{\textbf{Fig. \ref{fig:fig3} (b)}}-\textcolor{highlight}{\textbf{(c)}}. To unify the representation of nuclei at the semantic and structural levels, this paper develops a novel contour-based structure encoding to ensure semantic consistency inside or outside contours.

As shown in \textcolor{highlight}{\textbf{Fig. \ref{fig:fig3} (d)}}, the SE map ${y}_{str}$ encodes the integral structure of each nucleus. The element ${y}_{str}^{i,j}$ at $(i, j)$ location of ${y}_{str}$ represents a distance of the pixel $p^{i,j}$ to a specific nuclear contour. 
To differentiate the interior and exterior of nuclei, we use different signs for encoding ${y}_{str}^{i,j}$. 
Positive and negative values indicate whether the pixel is inside or outside the nucleus.
When $p^{i,j}$ lies on a contour, the encoding value ${y}_{str}^{i,j}$ is set to 0. %The above encoding procedure can be detailed as following steps.

Given a binarized semantic mask, the contour pixel set $\boldsymbol{\mathcal{C}}$ are derived from the mask through dilation and erosion operations. 
Afterward, the Euclidean distance is used to measure the value of the structure encoding. 
For a pixel within a nucleus, the $p_{inst}^{con}$ is used, which represents the contour pixel closest to the pixel within its instance, while for a pixel outside the nucleus, the nearest contour pixel $p_{near}^{con}$ is utilized as a reference. The contour-based structure encoding ${y}_{str}^{i,j}$ is formulated as:
\begin{equation}
	\begin{split}
		{y}_{str}^{i,j}= \left \{
		\begin{array}{ll}
			\Vert Loc(p^{i,j}) - Loc(p_{inst}^{con}) \Vert_2, & p^{i,j} \in \boldsymbol{\mathcal{I}}, \\
			- \Vert Loc(p^{i,j}) - Loc(p_{near}^{con})  \Vert_2, &  p^{i,j} \in \boldsymbol{\mathcal{O}}, \\
			0,    & p^{i,j} \in \boldsymbol{\mathcal{C}}, \\
		\end{array}
		\right.
	\end{split}
\end{equation}
where $\boldsymbol{\mathcal{I}}$ indicates the set of pixels inside the nuclei and $\boldsymbol{\mathcal{O}}$ represents the set of pixels outside the nuclei. $Loc(\cdot)$ represents the location of the pixel. A min-max normalization operation is used to transform the encoding values $y_{str}^{i,j}$ of each instance from $-1$ to $1$.

As illustrated in the \textcolor{highlight}{\textbf{Fig. \ref{fig:fig3} (d)}}, our structure encoding divides the nucleus into a series of isoheight lines, where each line denotes a type of similar structure. Meantime, the red area represents nuclei and the blue region represents the extracellular tissue. Notably, our modeling strategy unifies structure and semantic information, providing a bridge for structure interactions.

\begin{figure}[t]
	\centering
	\includegraphics[width=0.9\linewidth]{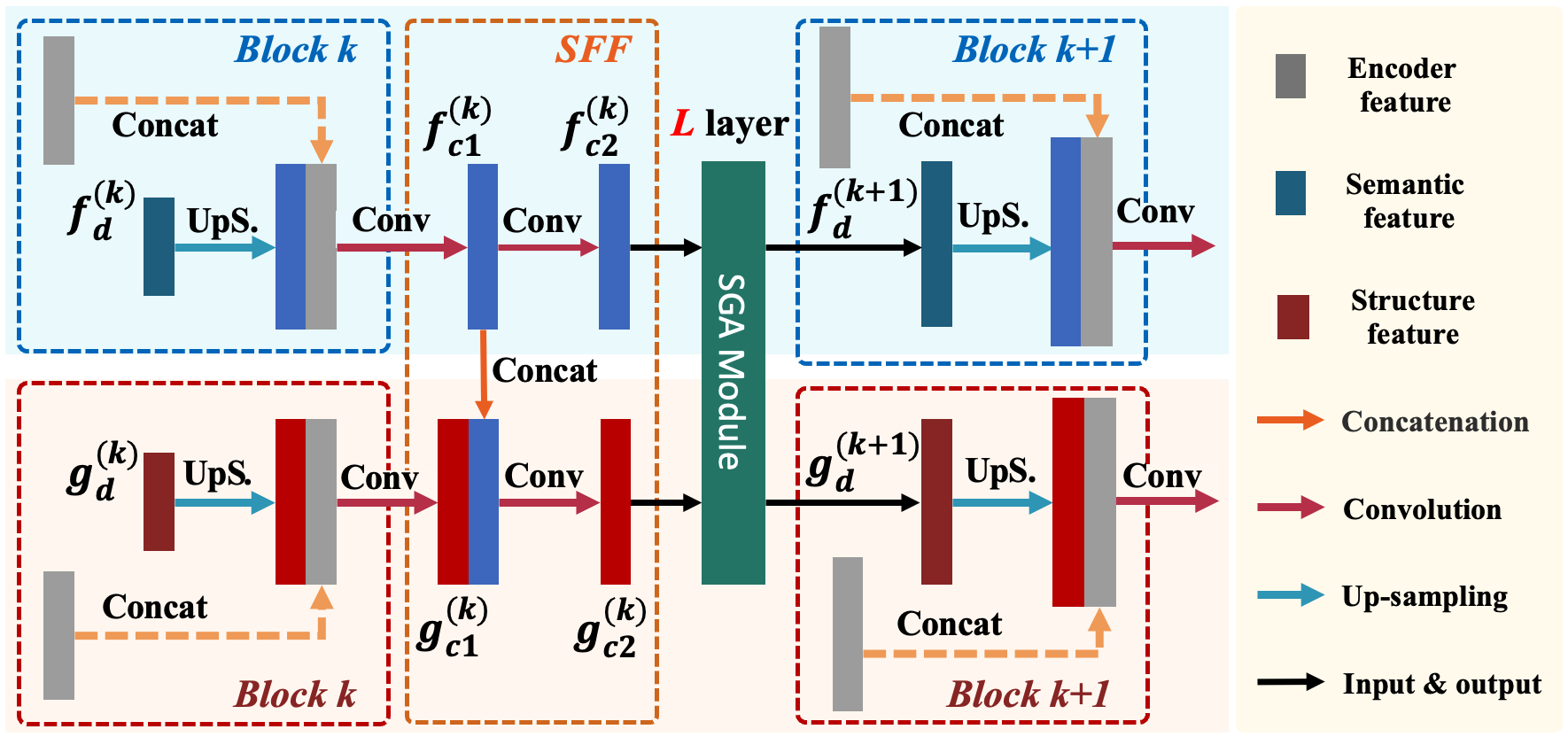}
	\caption{\textbf{The optimization procedure of the semantic feature and structure feature between two sequential blocks.} The feature outputs of SFF are fed into SGA for structure interactions.}
	\label{fig:fig4}
	\vspace{-0.3cm}
\end{figure}

\subsection{Semantic Feature Fusion}

To maintain semantic consistency of these branches, we propose a semantic feature fusion (SFF) that incorporate features of the semantic branch into the structure branch. The intuitive illustration is shown in \textcolor{highlight}{\textbf{Fig. \ref{fig:fig4}}}.

In \textcolor{highlight}{\textbf{Fig. \ref{fig:fig4}}}, the semantic feature $f_d^{(k)}$ is the output of the semantic branch $k$-th block, which is then upsampled and concatenated with the feature of encoder's $k$-th block. Afterwards, the concatenate feature is fed into a convolution layer for generating the enhanced semantic feature $f_{c1}^{(k)}$.
Similarily, we obtain the enhanced structure feature $g_{c1}^{(k)}$. To enhance the representation capability of the structure branch, we concatenate $f_{c1}^{(k)}$ and $g_{c1}^{(k)}$ for compensating semantic information for the structure feature. Next, $f_{c1}^{(k)}$ and $g_{c1}^{(k)}$ are respectively updated by a convolutional layer for generating the inputs of SGA:
\begin{equation}
	\begin{split}
	f_{c2}^{(k)}=Conv(f_{c1}^{(k)}), \ 
	g_{c2}^{(k)}=Conv(Concat(f_{c1}^{(k)},g_{c1}^{(k)})),
	\end{split}
\end{equation}
where $Conv$ represents the convolutional operation and $Concat$ represents the concatenation operation.

\subsection{Structure Guided Attention}
In histopathology image tiles, nuclei usually present similarities in the shape and size. The clearly stained nuclei have potentials for the network to comprehend the structure of chromophobe nuclei. For this goal, we introduce a structure-guided attention (SGA) mechanism for structure interaction between nuclei, thereby enhancing the structure learning of fuzzy nuclei and improving the feature representation of the semantic branch.

In SGA, three $1\times1$ convolution operations are respectively used to generate a query vector $\boldsymbol{{Q}_{str}} \in \mathbb{R}^{H \times W \times C}$, a key vector $\boldsymbol{{K}_{str}} \in \mathbb{R}^{H \times W \times C}$, and a value vector $\boldsymbol{{V}_{str}} \in \mathbb{R}^{H \times W \times C}$ from $g_{c2}^{(k)}$. 
Concurrently, a semantic value vector $\boldsymbol{V_{sem}} \in \mathbb{R}^{H \times W \times C}$ is derived from $f_{c2}^{(k)}$. Next, we flatten them to $\mathbb{R}^{HW\times C}$, and adopt a self-attention \cite{cai2022using, zhou2023nnformer, xu2022omega} operation to generate the structure-guided attention $\boldsymbol{S}$:
\begin{equation}
	\boldsymbol{S} = Softmax(\frac{\boldsymbol{{Q}_{str}}\boldsymbol{K_{str}^{T}}}{\sqrt{C}}),
\end{equation}
where $\boldsymbol{{K}_{str}^{T}}$ represents the transposed $\boldsymbol{{K}_{str}}$.

\begin{figure}[t]
	\centering
	\includegraphics[width=0.87\linewidth]{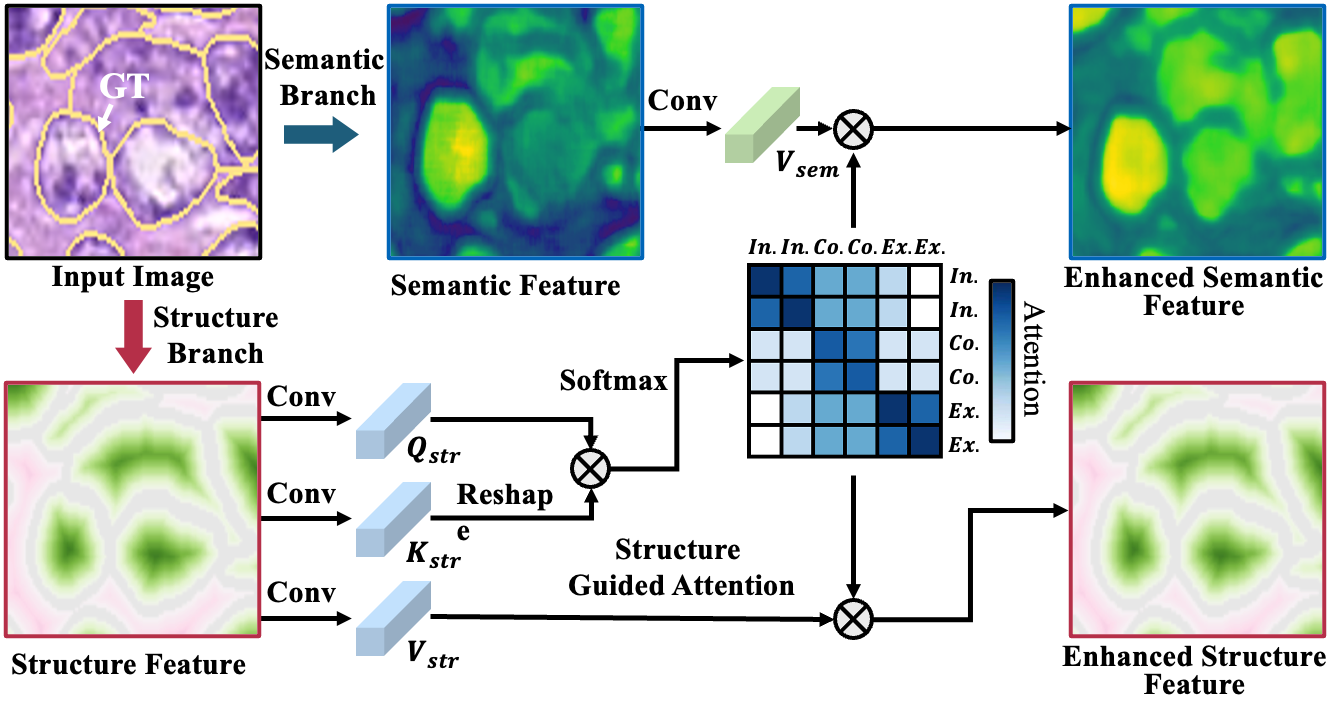}
	\caption{\textbf{The illustration of structure-guided attention module}. In the SGA map, the ``\textit{In.}", ``\textit{Co.}", and ``\textit{Ex.}" representents the intranuclear, contour, and extranuclear regions, respectively. }
	\label{fig:fig5}
	\vspace{-0.3cm}
\end{figure}

As shown in \textcolor{highlight}{\textbf{Fig. \ref{fig:fig5}}}, the attention map $\boldsymbol{S}$ highlights the structure relations between similar structure areas. Meanwhile, the attention assigns low scores for regions belonging to different structures, such as the intranuclear and extranuclear regions. Moreover, the structure correlation could reflect semantic relationships to some extents.

With the struture guided attention, we update structure and semantic features simultaneously, which can be formulated as:
\begin{equation}
	g_d^{(k+1)} = \boldsymbol{S} \cdot \boldsymbol{{V}_{str}}, \ 
	f_d^{(k+1)} = \boldsymbol{S} \cdot \boldsymbol{{V}_{sem}},
\end{equation}
where $g_d^{(k+1)}$ and $f_d^{(k+1)}$ represent the updated structure feature and semantic feature, respectively. Motivated by cross-criss attention \cite{huang2019ccnet}, we simplify the self-attention calculation as following steps. 
For each position $m$ in the spatial dimension of feature map $\boldsymbol{Q_{str}}$, we obtain a query vector $\boldsymbol{Q_m} \in \mathbb{R}^{C}$. Meanwhile, we extract the key vector set $\boldsymbol{\Omega_{m}} \in \mathbb{R}^{(H+W-1) \times C}$ from $\boldsymbol{K_{str}}$ with the same row or column of position $m$. Hence, the attention score $a_{m,n}$ between positions $m$ and $n$ can be calculated as follows:
\begin{equation}
	a_{m,n}=\boldsymbol{Q_{m}}\boldsymbol{\Omega_{m,n}^{T}},
\end{equation}
where $\boldsymbol{\Omega_{m,n}} \in \mathbb{R}^{C}$ is the $n$-th element of $\boldsymbol{\Omega_{m}}$. $a_{m,n} \in \boldsymbol{A}$ is the degree of correlation between feature $\boldsymbol{Q_m}$ and $\boldsymbol{\Omega_{m,n}}$. Then, we apply a softmax operation on $\boldsymbol{A}$ over the channel dimension to calculate the criss-cross attention map. In the SGA module, we stack the operation twice to obtain the updated strcuture feature $g_d^{(k+1)}$ and semantic feature $f_d^{(k+1)}$.

\begin{figure}[t]
	\centering
	\includegraphics[width=0.9\linewidth]{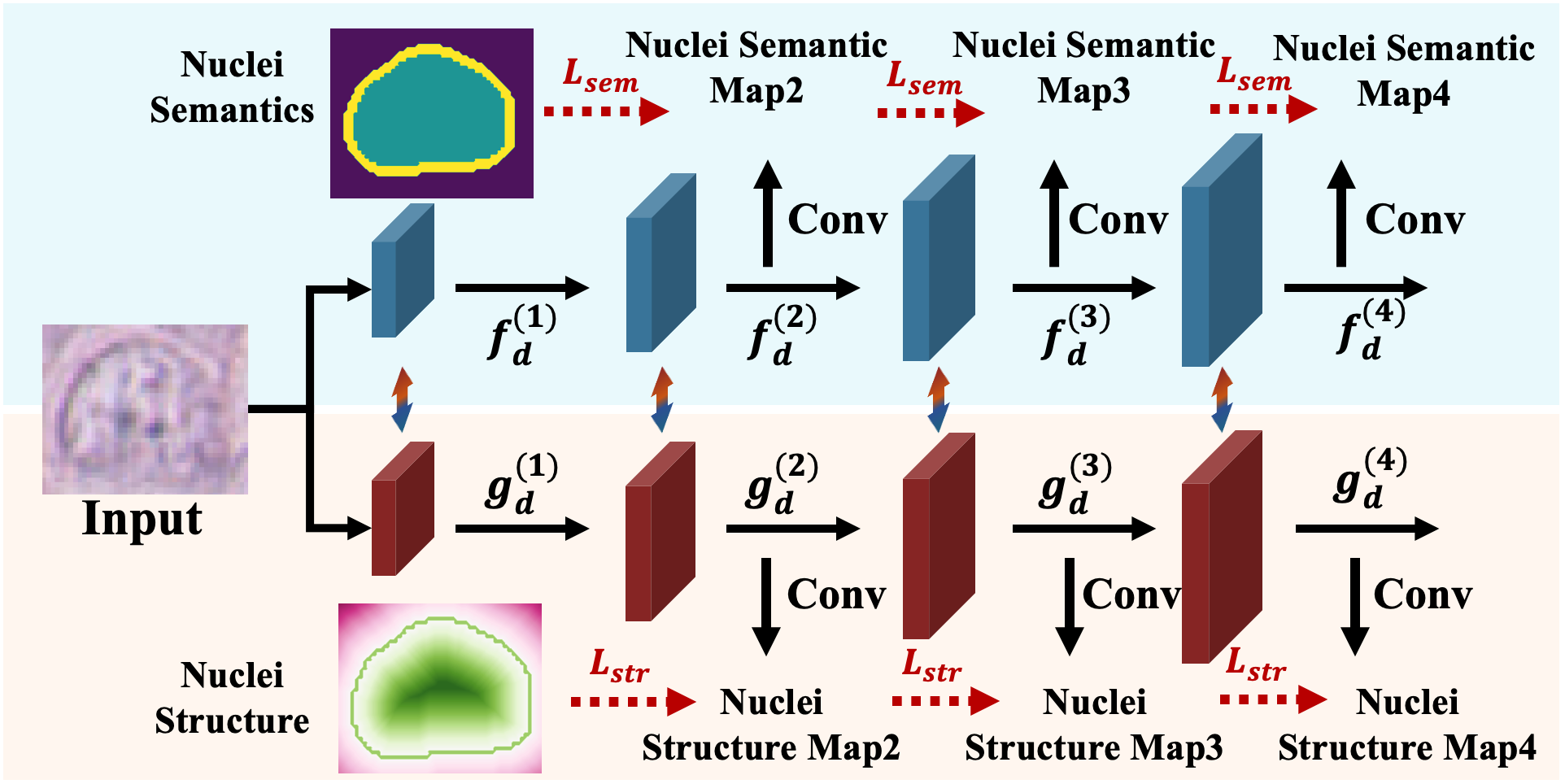}
	\caption{\textbf{The multi-scale optimization objective of semantic and structure branches.} The double-headed arrow represents the feature interaction between the two branches.}
	\label{fig:fig6}
	\vspace{-0.3cm}
\end{figure}

Throughout the training process of the network, we employ a multi-scale loss to optimize the semantic and structure branches as shown in \textcolor{highlight}{\textbf{Fig. \ref{fig:fig6}}}. In this process, we only calculate the semantic and structural losses on the feature maps from the second to the fourth blocks, since smaller-sized feature maps in the earlier blocks may fail to capture the intricate structural features of nuclei. 

SGA is performed at multiple levels in the decoder stage, finally generating a semantic feature $f_d^{(k)}$ and a structure feature $g_d^{(k)}$. The former is then fed into a convolutional layer to generate the semantic segmentation result $\hat{y}_{sem}^{(k)}$ of the $k$-th block, which is optimzed by a semantic segmentation loss $L_{sem}$:
\begin{equation}\small{
		\begin{split}
			L_{sem} &= \sum_{k=2}^{4} \{CE(\hat{y}_{sem}^{(k)}, y_{sem}^{(k)}) + Dice(\hat{y}_{sem}^{(k)}, y_{sem}^{(k)})\},
	\end{split}}
\end{equation}
where $CE$ and $Dice$ represent cross-entropy loss and Dice loss \cite{abdollahi2020vnet}. 
Similarly, with the $1 \times 1$ convolutional layer for $g_d^{(k)}$, we optimize the structure branch prediction $\hat{y}_{str}$ through a structure loss $L_{str}$:
\begin{equation}
	L_{str}= \sum_{k=2}^{4} \{MSE(\hat{y}_{str}^{(k)}-y_{str}^{(k)})\},
\end{equation}
where MSE represents the mean squared error. $y_{sem}^{(4)}$ and $y_{str}^{(4)}$ are the ground truths (GTs) of the semantic and structure branches. Low scale GTs are derived by downsampling to $y_{sem}^{(4)}$ and $y_{str}^{(4)}$.

\subsection{Position Enhancement}
In the structure modeling mechanism, the encoding takes contours as references. However, the contour-focus task may be disturbed by complex textures in intranuclear regions, leading to wrong instance segmentation results, such as an individual nucleus being segmented into multiple nuclei. 
To maintain structure integrality of nuclei, we propose a position enhancement (PE) offering prior location knowledge to suppress the wrong boundary predictions. 

Specifically, PE consists of a residual unit and an $1\times1$ convolutional layer as shown in the orange box of \textcolor{highlight}{\textbf{Fig. \ref{fig:fig2}}}.
We firstly feed features $f_d^{(4)}$ and $g_d^{(4)}$ into the residual unit respectively. These procedures reduce the conflict between semantic (or structure) and position prediction tasks. The output of the residual unit is fed into the $1 \times 1$ convolutional layer to predict the distance from each pixel to centroid. The distance can be formulated as follows:
\begin{equation}
		\begin{split}
		{y}_{pos}^{i,j}= \left \{
		\begin{array}{ll}
			\Vert Loc(p^{i,j}) - Loc(p^{cen}) \Vert_2, & p^{i,j} \in \boldsymbol{\mathcal{I}} \cup \boldsymbol{\mathcal{C}}, \\
			0,    & p^{i,j} \in \boldsymbol{\mathcal{O}}, \\
		\end{array}
		\right.
	\end{split}
\end{equation}
where $p^{cen}$ is the centroid of nuclei. The loss function of the two PE modules is following:
\begin{equation}
	L_{pos} = MSE(\hat{y}_{pos}^{sem}-y_{pos}) + MSE(\hat{y}_{pos}^{str}-y_{pos}),
\end{equation}
where the $\hat{y}_{pos}^{sem}$ and $\hat{y}_{pos}^{str}$ are the predicted position map of semantic branch and structure branch, respectively.

\subsection{Post Processing}

With the structure encoding, we can obtain high-confidence contours, which are beneficial for distinguishing nuclei boundary. As shown in bottom of \textbf{\textcolor{highlight}{Fig. \ref{fig:fig2}}}, post processing is used to fuse the structure prediction into the semantic prediction. Specifically, we input the image into the SEINE, obtaining a three-class mask map and a structure encoding map. With the two results, we apply two threshold values $t_p$ and $t_n$ to filter the structure encoding, generating a high-confidence contour map $\hat{y}_{con}$:
\begin{equation}
	\hat{y}_{con}= B(t_n <\hat{y}_{str}<t_p),
\end{equation}
where $B(\cdot)$ represents a binariy operation, $t_p - t_n$ represents the relative width of the generated contours. In this paper, we set $t_p=0.05$ and $t_n=-0.05$. Afterwards, we fuse the semantic map and the contour map to generate a refined mask, which serves as the final segmentation result.

\section{Experiments}\label{exper}

\definecolor{green}{RGB}{0,0,0}
\definecolor{red}{RGB}{0,0,0}
\definecolor{blue}{RGB}{0,0,0}
\begin{table*}[t]
	\caption{\textbf{The ablation experiments of stucture encoding methods on ConSeP, MoNuSeg, PanNuke, and TNBC datasets.} ``SE" represents our proposed structure encoding. ``+SGA" represents the introduction of SGA module.}
	\vspace{-0.4cm}
	\begin{center}
		\renewcommand{\arraystretch}{1.3}
		\resizebox{\linewidth}{!}{
			\begin{tabular}{c|l|llll|llll|llll|llll}
				\Xhline{1.4pt}
				\multirow{2}{*}{\textbf{Methods}} &  \textbf{Datasets} &  \multicolumn{4}{c|}{\textbf{ConSeP}} & \multicolumn{4}{c}{\textbf{MoNuSeg}}& \multicolumn{4}{c|}{\textbf{PanNuke}} & \multicolumn{4}{c}{\textbf{TNBC}} \\ 
				\cline{2-18}
				& \textbf{Metrics} & \textbf{Dice(\%) $\uparrow$} &\textbf{AJI(\%)$\uparrow$} & \textbf{HausD$\downarrow$} & \textbf{PQ(\%)$\uparrow$} & \textbf{Dice(\%)$\uparrow$} & \textbf{AJI(\%)$\uparrow$} & \textbf{HausD$\downarrow$} & \textbf{PQ(\%)$\uparrow$}& \textbf{Dice(\%)$\uparrow$ }& \textbf{AJI(\%)$\uparrow$} & \textbf{HausD$\downarrow$} & \textbf{PQ(\%)$\uparrow$} & \textbf{Dice(\%)$\uparrow$} & \textbf{AJI(\%)$\uparrow$} & \textbf{HausD$\downarrow$} & \textbf{PQ(\%)$\uparrow$} \\ 
				\Xhline{1.4pt}
				& \textbf{w contour} &78.6& 51.6 & 12.97 & 43.3 & 79.5& 56.5 & \underline{8.69} & 51.0 & 79.8 & 58.4 & 9.44 & 52.4 & 76.6 & 61.4 & 6.97 & 57.7  \\
				& \textcolor{blue}{\textbf{HV}} & \textcolor{blue}{{79.1}$_{+0.5}$} & \textcolor{blue}{{53.1}$_{+1.5}$} & \textcolor{blue}{{11.77}$_{-1.20}$} &\textcolor{blue}{{43.6}$_{+0.3}$} & \textcolor{blue}{{77.8}$_{-1.7}$} & \textcolor{blue}{{56.4}$_{-0.1}$} & \textcolor{blue}{{11.23}$_{+2.54}$} & \textcolor{blue}{{50.0}$_{-1.0}$} & \textcolor{blue}{\underline{81.0}$_{+1.2}$} & \textcolor{blue}{{61.8}$_{+3.4}$} & \textcolor{blue}{{7.20}$_{-2.24}$} & \textcolor{blue}{{56.4}$_{+4.0}$} & \textcolor{blue}{{76.2}$_{-0.4}$} & \textcolor{blue}{{66.3}$_{+4.9}$}& \textcolor{blue}{{5.24}$_{-1.73}$} & \textcolor{blue}{{58.3}$_{+0.6}$}  \\
				& \textcolor{green}{\textbf{Dir}} & \textcolor{green}{75.6$_{-3.0}$} & \textcolor{green}{47.6$_{-4.0}$}  &  \textcolor{green}{14.81$_{+1.84}$}  & \textcolor{green}{41.2$_{-2.1}$} & \textcolor{green}{72.8$_{-6.7}$} & \textcolor{green}{55.7$_{-0.8}$} & \textcolor{green}{13.96$_{+5.27}$} & \textcolor{green}{49.3$_{-1.7}$} & \textcolor{green}{78.9$_{-0.9}$} & \textcolor{green}{61.1$_{+2.7}$} & \textcolor{green}{8.96$_{+0.48}$} & \textcolor{green}{53.6$_{+1.2}$} & \textcolor{green}{77.8$_{+1.2}$} & \textcolor{green}{65.4$_{+4.0}$} & \textcolor{green}{5.59$_{-1.38}$} & \textcolor{green}{58.6$_{+0.9}$}  \\
				\textbf{UNet} & \textcolor{red}{\textbf{SE}} &\textcolor{red}{\underline{80.2}$_{+1.6}$}  & \textcolor{red}{\underline{53.3}$_{+1.7}$}  & \textcolor{red}{11.69$_{-1.28}$} & \textcolor{red}{\underline{44.4}$_{+1.1}$} & \textcolor{red}{80.3$_{+0.8}$} & \textcolor{red}{\underline{57.6}$_{+1.1}$} & \textcolor{red}{8.75$_{+0.06}$} & \textcolor{red}{51.3$_{+0.3}$} &\textcolor{red}{80.4$_{+0.6}$} & \textcolor{red}{62.1$_{+3.7}$} & \textcolor{red}{7.11$_{-2.33}$} & \textcolor{red}{56.7$_{+4.3}$} & \textcolor{red}{78.3$_{+1.7}$} & \textcolor{red}{66.8$_{+5.4}$} & \textcolor{red}{5.14$_{-1.83}$} & \textcolor{red}{58.5$_{+0.8}$}  \\
				& \textcolor{blue}{\textbf{HV+SGA}} & \textcolor{blue}{79.5$_{+0.4}$} & \textcolor{blue}{53.1$_{+0.0}$} & \textcolor{blue}{\underline{11.08}$_{-0.69}$} & \textcolor{blue}{43.9$_{+0.3}$} &  \textcolor{blue}{\underline{81.2}$_{+1.7}$} & \textcolor{blue}{57.0$_{+0.5}$} & \textcolor{blue}{6.33$_{-2.33}$} & \textcolor{blue}{\underline{52.6}$_{+1.6}$} & \textcolor{blue}{80.3$_{+0.5}$} & \textcolor{blue}{\underline{64.3}$_{+5.9}$} &  \textcolor{blue}{\underline{6.88}$_{-2.56}$} & \textcolor{blue}{\underline{57.2}$_{+4.8}$} & \textcolor{blue}{\underline{78.4}$_{+1.8}$} & \textcolor{blue}{\underline{67.3}$_{+5.9}$} & \textcolor{blue}{\underline{5.07}$_{-1.90}$} & \textcolor{blue}{\underline{59.7}$_{+2.0}$} \\
				& \textcolor{green}{\textbf{Dir+SGA}} & \textcolor{green}{78.3$_{+2.7}$} & \textcolor{green}{48.4$_{+0.8}$} & \textcolor{green}{14.70$_{-0.11}$} & \textcolor{green}{38.5$_{-2.7}$} & \textcolor{green}{80.5$_{+7.7}$} & \textcolor{green}{50.0$_{-5.7}$} & \textcolor{green}{16.71$_{+2.75}$} & \textcolor{green}{48.1$_{-1.2}$} & \textcolor{green}{79.4$_{+0.5}$}& \textcolor{green}{61.3$_{+0.2}$}&  \textcolor{green}{8.04$_{-0.92}$} & \textcolor{green}{54.2$_{+0.6}$}& \textcolor{green}{78.3$_{+0.5}$} & \textcolor{green}{66.0$_{+0.6}$} &  \textcolor{green}{5.43$_{-0.16}$} & \textcolor{green}{58.9$_{+0.3}$} \\
				& \textcolor{red}{\textbf{SE+SGA}} & \textcolor{red}{\textbf{81.5$_{+1.3}$}} & \textcolor{red}{\textbf{53.6$_{+0.3}$}} & \textcolor{red}{\textbf{9.21$_{-2.48}$}} & \textcolor{red}{\textbf{46.5$_{+2.1}$}} & \textcolor{red}{\textbf{83.6$_{+3.3}$}} & \textcolor{red}{\textbf{60.3$_{+2.7}$}} & \textcolor{red}{\textbf{6.12$_{-2.63}$}} & \textcolor{red}{\textbf{57.4$_{+6.1}$}} & \textcolor{red}{\textbf{82.7$_{+2.3}$}} & \textcolor{red}{\textbf{67.0$_{+4.9}$}} & \textcolor{red}{\textbf{4.86$_{-2.25}$}} & \textcolor{red}{\textbf{61.1$_{+4.4}$}} & \textcolor{red}{\textbf{78.9$_{+0.6}$}} & \textcolor{red}{\textbf{67.9$_{+1.1}$}} & \textcolor{red}{\textbf{4.98$_{-0.16}$}} & \textcolor{red}{\textbf{63.6$_{+5.1}$}} \\
				\Xhline{0.8pt}
				& \textbf{w contour} & 81.2 &54.8 & 8.87 &48.7& 80.3 & 56.7 & 8.74 & 50.9 & 80.4 & 59.3 & 9.12 & 53.7 & 77.9 & 63.8 & 6.02 & 58.6  \\
				& \textcolor{blue}{\textbf{HV}} & \textcolor{blue}{81.6$_{+0.4}$} & \textcolor{blue}{54.3$_{-0.5}$} & \textcolor{blue}{9.79$_{+0.92}$} &  \textcolor{blue}{48.2$_{-0.5}$} & \textcolor{blue}{81.7$_{+1.4}$} & \textcolor{blue}{57.4$_{+0.7}$} & \textcolor{blue}{8.09$_{-0.65}$} & \textcolor{blue}{52.3$_{+1.4}$} & \textcolor{blue}{80.3$_{+0.1}$} & \textcolor{blue}{62.1$_{+2.8}$} & \textcolor{blue}{7.47$_{-1.65}$} & \textcolor{blue}{55.2$_{+1.5}$} & \textcolor{blue}{76.7$_{-1.2}$} & \textcolor{blue}{65.6$_{+1.8}$}  & \textcolor{blue}{5.49$_{-0.53}$} & \textcolor{blue}{60.0$_{+1.4}$}  \\
				& \textcolor{green}{\textbf{Dir}} & \textcolor{green}{81.4$_{+0.2}$} & \textcolor{green}{53.3$_{-1.5}$} & \textcolor{green}{12.30$_{+3.43}$} & \textcolor{green}{47.9$_{-0.8}$} & \textcolor{green}{81.5$_{+1.2}$} & \textcolor{green}{56.5$_{-0.2}$} & \textcolor{green}{8.98$_{+0.24}$} & \textcolor{green}{51.8$_{+0.9}$} & \textcolor{green}{79.7$_{-0.7}$} & \textcolor{green}{61.1$_{+1.8}$} & \textcolor{green}{8.98$_{-0.14}$} & \textcolor{green}{54.0$_{+0.3}$} & \textcolor{green}{76.4$_{-1.5}$} & \textcolor{green}{65.3$_{+1.5}$} & \textcolor{green}{5.81$_{-0.21}$} & \textcolor{green}{59.1$_{+0.5}$}  \\
				\textbf{MicroNet} & \textcolor{red}{\textbf{SE}} &  \textcolor{red}{\underline{81.8}$_{+0.6}$} &  \textcolor{red}{\underline{55.0}$_{+0.2}$} &  \textcolor{red}{\underline{8.20}$_{-0.67}$} &  \textcolor{red}{\underline{48.9}$_{+0.2}$} &  \textcolor{red}{\underline{82.3}$_{+2.0}$} &  \textcolor{red}{58.8$_{+2.1}$} &  \textcolor{red}{6.53$_{-2.21}$} &  \textcolor{red}{54.7$_{+3.8}$} &  \textcolor{red}{\underline{80.8}$_{+0.4}$} &  \textcolor{red}{64.4$_{+5.1}$} &  \textcolor{red}{\underline{6.50}$_{-2.62}$} &  \textcolor{red}{56.9$_{+3.2}$} &  \textcolor{red}{77.8$_{-0.1}$} &  \textcolor{red}{66.2$_{+2.4}$} &  \textcolor{red}{5.12$_{-0.90}$} &  \textcolor{red}{60.5$_{+1.9}$}  \\
				%				\cline{2-18}
				& \textcolor{blue}{\textbf{HV+SGA}} & \textcolor{blue}{81.4$_{+0.2}$} & \textcolor{blue}{54.9$_{+0.6}$} & \textcolor{blue}{8.54$_{-1.25}$} & \textcolor{blue}{47.7$_{-0.5}$} & \textcolor{blue}{82.2$_{+0.5}$} & \textcolor{blue}{58.3$_{+0.9}$} & \textcolor{blue}{6.71$_{-1.38}$} & \textcolor{blue}{\underline{54.9}$_{+2.6}$} & \textcolor{blue}{80.2$_{-0.1}$} & \textcolor{blue}{\underline{65.0}$_{+2.9}$} & \textcolor{blue}{6.93$_{-0.54}$} & \textcolor{blue}{\underline{58.8}$_{+3.6}$} &  \textcolor{blue}{\textbf{79.4}$_{+2.7}$} & \textcolor{blue}{\underline{68.1}$_{+.43}$} & \textcolor{blue}{\underline{4.79}$_{-1.23}$} & \textcolor{blue}{\underline{61.9}$_{+3.3}$} \\
				& \textcolor{green}{\textbf{Dir+SGA}} & \textcolor{green}{79.1$_{-2.3}$} & \textcolor{green}{51.7$_{-1.6}$} & \textcolor{green}{13.22$_{-0.92}$} & \textcolor{green}{44.5$_{-3.4}$} & \textcolor{green}{82.0$_{+0.5}$} & \textcolor{green}{\underline{59.6}$_{+3.1}$} & \textcolor{green}{\underline{5.84}$_{-3.15}$} & \textcolor{green}{53.7$_{+1.9}$} & \textcolor{green}{80.1$_{+0.4}$} & \textcolor{green}{64.8$_{+3.7}$} & \textcolor{green}{6.88$_{-2.10}$} & \textcolor{green}{57.6$_{+3.6}$} & \textcolor{green}{\underline{78.9}$_{+2.5}$} & \textcolor{green}{67.9$_{+2.6}$} & \textcolor{green}{4.85$_{-0.96}$} & \textcolor{green}{61.6$_{+2.5}$} \\
				& \textcolor{red}{\textbf{SE+SGA}} &  \textcolor{red}{\textbf{82.4}$_{+0.6}$} &  \textcolor{red}{\textbf{55.6}$_{+0.6}$} &  \textcolor{red}{\textbf{6.97}$_{-1.23}$} &  \textcolor{red}{\textbf{50.3}$_{+1.4}$} &  \textcolor{red}{\textbf{83.1}$_{+0.8}$} &  \textcolor{red}{\textbf{60.9}$_{+2.1}$} &  \textcolor{red}{\textbf{5.17}$_{-1.36}$} &  \textcolor{red}{\textbf{56.1}$_{+1.4}$} &  \textcolor{red}{\textbf{81.3}$_{+0.5}$} &  \textcolor{red}{\textbf{65.5}$_{+1.1}$} &  \textcolor{red}{\textbf{6.06}$_{-0.44}$} &  \textcolor{red}{\textbf{59.8}$_{+2.9}$} &  \textcolor{red}{78.7$_{+0.9}$} &  \textcolor{red}{\textbf{68.2}$_{+2.0}$} &  \textcolor{red}{\textbf{4.64}$_{-0.48}$} &  \textcolor{red}{\textbf{62.0}$_{+1.5}$}  \\
				\Xhline{1.0pt}
				& \textbf{w contour} &84.0 &51.9 & 11.20 &42.6 & 80.4 & 57.7 & 9.21 & 50.5 & 81.3 & 62.2 & 8.49 & 53.4 &78.1 & 65.4 & 6.73 & 57.9  \\
				&  \textcolor{blue}{\textbf{HV}} & \textcolor{blue}{83.4$_{-0.6}$} & \textcolor{blue}{51.7$_{-0.2}$} & \textcolor{blue}{10.88$_{-0.32}$} & \textcolor{blue}{43.1$_{+0.5}$} & \textcolor{blue}{80.7$_{+0.3}$} & \textcolor{blue}{59.5$_{+1.8}$} &  \textcolor{blue}{7.73$_{-1.48}$} & \textcolor{blue}{50.7$_{+0.2}$} & \textcolor{blue}{82.0$_{+0.7}$} & \textcolor{blue}{62.8$_{+0.6}$} & \textcolor{blue}{7.97$_{-0.52}$} & \textcolor{blue}{53.7$_{+0.3}$} & \textcolor{blue}{79.6$_{+1.5}$} & \textcolor{blue}{66.2$_{+0.8}$} & \textcolor{blue}{6.41$_{-0.32}$} & \textcolor{blue}{57.7$_{-0.2}$}  \\
				&\textcolor{green}{\textbf{Dir}} & \textcolor{green}{83.9$_{-0.1}$} & \textcolor{green}{52.1$_{+0.2}$} & \textcolor{green}{9.67$_{-1.53}$} & \textcolor{green}{43.5$_{+0.9}$} & \textcolor{green}{79.4$_{-1.0}$} &\textcolor{green}{57.5$_{-0.2}$} & \textcolor{green}{9.89$_{+0.68}$} & \textcolor{green}{50.5$_{+0.0}$} & \textcolor{green}{81.8$_{+0.5}$} &\textcolor{green}{63.6$_{+1.4}$} & \textcolor{green}{7.07$_{-1.42}$} & \textcolor{green}{52.9$_{-0.5}$} & \textcolor{green}{79.1$_{+1.0}$} & \textcolor{green}{66.1$_{+0.7}$} & \textcolor{green}{6.55$_{-0.18}$} & \textcolor{green}{57.3$_{-0.6}$}  \\
				\textbf{Dist} &  \textcolor{red}{\textbf{SE}} &  \textcolor{red}{\underline{84.3}$_{-0.3}$} &  \textcolor{red}{\underline{52.3}$_{+0.4}$} &  \textcolor{red}{\underline{9.06}$_{-2.14}$} &  \textcolor{red}{43.6$_{+1.0}$} &  \textcolor{red}{\underline{83.2}$_{+2.8}$} &  \textcolor{red}{\underline{59.8}$_{+2.1}$} &  \textcolor{red}{6.21}$_{-3.00}$ &  \textcolor{red}{51.2$_{+0.7}$} &  \textcolor{red}{\underline{82.3}$_{+1.0}$} &  \textcolor{red}{\underline{63.9}$_{+1.7}$} &  \textcolor{red}{\underline{6.76}$_{-1.73}$} &  \textcolor{red}{\underline{55.2}$_{+1.8}$} &  \textcolor{red}{\underline{79.7}$_{+1.6}$} &  \textcolor{red}{66.2$_{+0.8}$} &  \textcolor{red}{6.38$_{-0.35}$} &  \textcolor{red}{57.6$_{-0.3}$}  \\
				&\textcolor{blue}{\textbf{HV+SGA}} & \textcolor{blue}{83.6$_{-0.4}$} & \textcolor{blue}{52.0$_{+0.3}$} & \textcolor{blue}{9.71$_{-1.17}$} & \textcolor{blue}{43.4$_{+0.3}$}  & \textcolor{blue}{81.2$_{+0.5}$} & \textcolor{blue}{\underline{59.8}$_{+0.3}$} & \textcolor{blue}{\underline{5.97}$_{+1.76}$} & \textcolor{blue}{\textbf{51.7}$_{+1.0}$} & \textcolor{blue}{81.8$_{-0.2}$} & \textcolor{blue}{62.6$_{-0.2}$} & \textcolor{blue}{8.21$_{+0.24}$} & \textcolor{blue}{54.0$_{+0.3}$} & \textcolor{blue}{\underline{79.7}$_{+0.1}$} & \textcolor{blue}{\underline{66.5}$_{+0.3}$} & \textcolor{blue}{6.20$_{-0.21}$} & \textcolor{blue}{57.4$_{-0.3}$}  \\
				& \textcolor{green}{\textbf{Dir+SGA}} & \textcolor{green}{79.1$_{-4.8}$} & \textcolor{green}{51.7$_{-0.4}$} &  \textcolor{green}{10.78$_{-0.10}$} & \textcolor{green}{\textbf{44.5}$_{+1.0}$} & \textcolor{green}{79.6$_{+0.2}$} & \textcolor{green}{58.0$_{+0.5}$} & \textcolor{green}{8.74$_{-1.15}$} & \textcolor{green}{50.8$_{+0.3}$} & \textcolor{green}{81.6$_{-0.2}$} & \textcolor{green}{62.9$_{-0.7}$} & \textcolor{green}{8.04$_{+0.97}$} & \textcolor{green}{53.0$_{+0.1}$} & \textcolor{green}{79.3$_{+0.2}$} & \textcolor{green}{65.9$_{-0.2}$} & \textcolor{green}{\underline{6.07}$_{-0.48}$} & \textcolor{green}{\underline{58.1}$_{+0.8}$}  \\
				& \textcolor{red}{\textbf{SE+SGA}} &  \textcolor{red}{\textbf{84.6}$_{+0.6}$} &  \textcolor{red}{\textbf{52.6}$_{+0.3}$} & \textcolor{red}{\textbf{8.34}$_{-0.72}$} &  \textcolor{red}{\underline{43.7}$_{+0.1}$} &  \textcolor{red}{\textbf{83.5}$_{+0.3}$} &  \textcolor{red}{\textbf{60.4}$_{+0.6}$} &  \textcolor{red}{\textbf{5.45}$_{-0.76}$} &  \textcolor{red}{\underline{51.4}$_{+0.2}$} &  \textcolor{red}{\textbf{82.5}$_{+0.2}$} &  \textcolor{red}{\textbf{64.3}$_{+0.4}$} &  \textcolor{red}{\textbf{6.11}$_{-0.65}$}  &  \textcolor{red}{\textbf{56.8}$_{+1.6}$} &  \textcolor{red}{\textbf{79.8}$_{+0.1}$} &  \textcolor{red}{\textbf{67.4}$_{+1.2}$} &  \textcolor{red}{\textbf{5.84}$_{-0.54}$} &  \textcolor{red}{\textbf{59.3}$_{+1.7}$}  \\
				\Xhline{1.4pt}
		\end{tabular}}
		\label{tab:ab1}
	\end{center}
	\vspace{-0.6cm}
\end{table*}

\subsection{Datasets and Evaluation Metrics}
We assess our method on four popular nuclei segmentation datasets, namely, ConSeP \cite{graham2019hover}, MoNuSeg \cite{kumar2017dataset}, PanNuke \cite{gamper2020pannuke}, and TNBC \cite{naylor2018segmentation}.
\subsubsection{ConSeP Dataset}
The colorectal nuclei segmentation dataset collected from UHCW (University Hospital Coventry and Warwickshire), which contains 41 hematoxylin-eosin (H\&E) stained histopathology image tiles of size 1000 × 1000.% with a total of 24,319 annotated nuclei.

\subsubsection{MoNuSeg Dataset} 
The multi-organ nuclei segmentation dataset was collected from TCGA (The Cancer Genome Atlas), and the data contains 30 H\&E stained histopathology image tiles of size 1000 × 1000 from 7 different organs.% with a total of 21,623 annotated nuclei.

\subsubsection{PanNuke Dataset}
The pan-cancer nuclei segmentation dataset collected from TCGA, contains more than 2,000 visual fields of size 256 × 256 in 19 different tissues. % with a total of 205,343 annotated nuclei.

\subsubsection{TNBC Dataset}
The triple-negative breast cancer dataset comes from Curie Institute, which consists of 50 H\&E stained histopathology image tiles of size 512 × 512.% from 11 breast cancer patients. 

%Refering to previous studies \cite{graham2019hover}, we preprocess each image into 256 $\times$ 256 size with a 128 step size. After splitting the images, we obtain 1323 training patches, 343 validation patches, and 336 testing patches on the ConSeP dataset. On the MoNuSeg dataset, we collect 784 training patches, 392 validation patches, and 294 testing patches. On the PanNuke dataset, we collect 2656 training patches, 2523 validation patches, and 2722 testing patches. On the TNBC dataset, we collect 296 training patches, 90 validation patches, and 64 testing patches. 
To precisely evaluate the effectiveness of our method, we adopt four widely-used metrics including Dice \cite{vu2019methods}, aggregated jaccard index (AJI) \cite{kumar2017dataset}, Hausdorff distance \cite{beauchemin1998hausdorff} and panoptic quality (PQ) \cite{kirillov2019panoptic}.

\definecolor{green}{RGB}{0,212,65}
\definecolor{red}{RGB}{255,0,0}
\definecolor{blue}{RGB}{0,199,255}
\begin{figure}[t]
	\centering
	\includegraphics[width=0.94\linewidth]{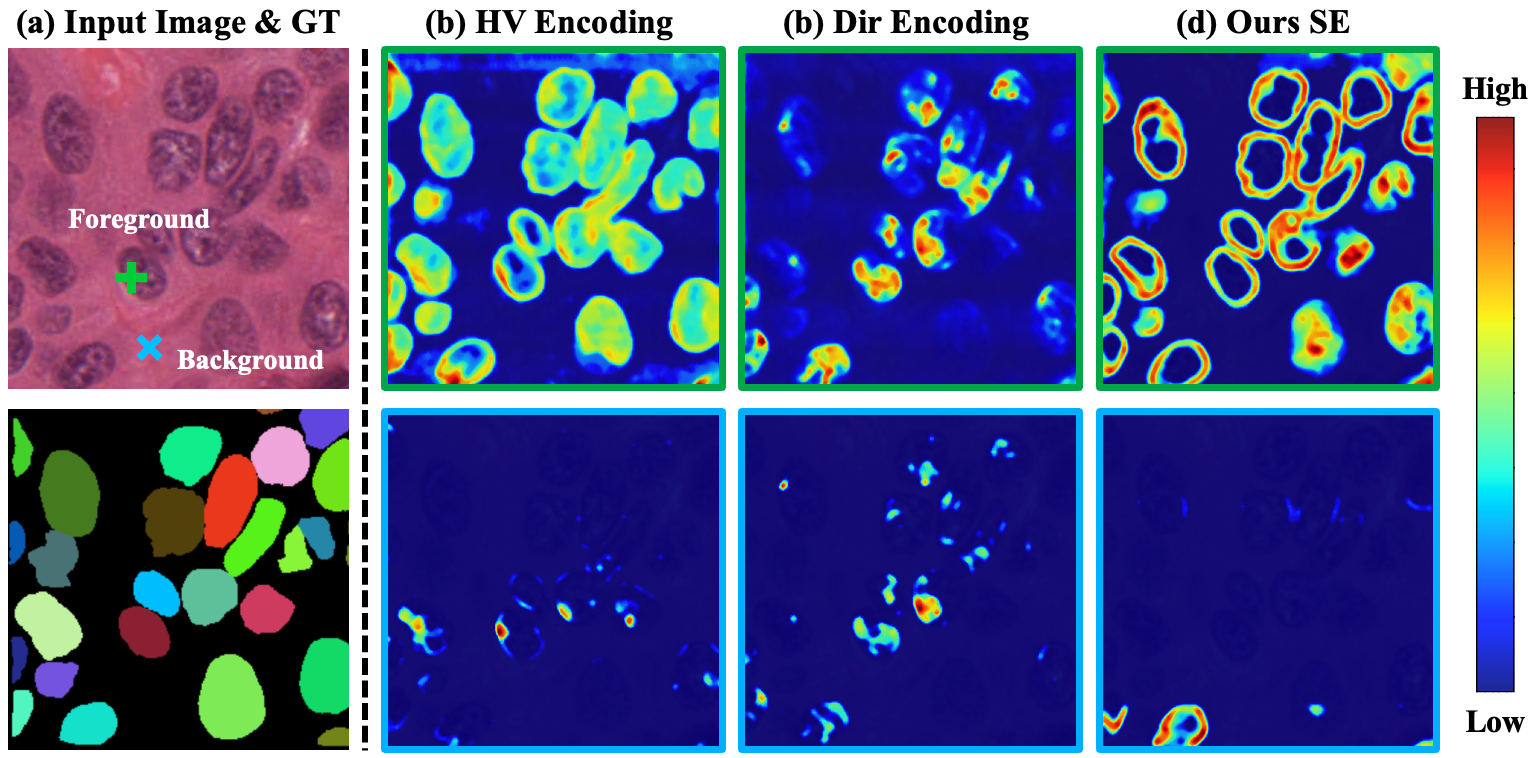}
	\caption{\textbf{The attention maps based on different structure modeling strategies.} The first row and the second row represent the attention maps generated based on a foreground pixel (\textbf{\textcolor{green}{+}}) and a background pixel (\textbf{\textcolor{blue}{×}}), respectively.}
	\label{fig:fig9}
	\vspace{-0.3cm}
\end{figure}
\subsection{Implementation Details }
Our experiments are implemented on PyTorch 1.10.1 with an Nvidia RTX 3090 GPU. We adopt a ResNet-50 \cite{he2016deep} as the feature extractor and use UNet decoder \cite{ronneberger2015u} to build the semantic branch and structure branch. Adam algorithm is employed to optimize our network, in which $\beta_1$ is set to 0.9, $\beta_2$ is set to 0.999, and the learning rate is set to 0.0005. Besides, the total epoch of training and the weight decay are set to 100 and 0.0005, respectively. 
In the training phase, we apply online data augmentation, including random rotation,  flipping, and scaling, to enhance the robustness of the model.

\begin{figure}[t]
	\centering
	\includegraphics[width=0.95\linewidth]{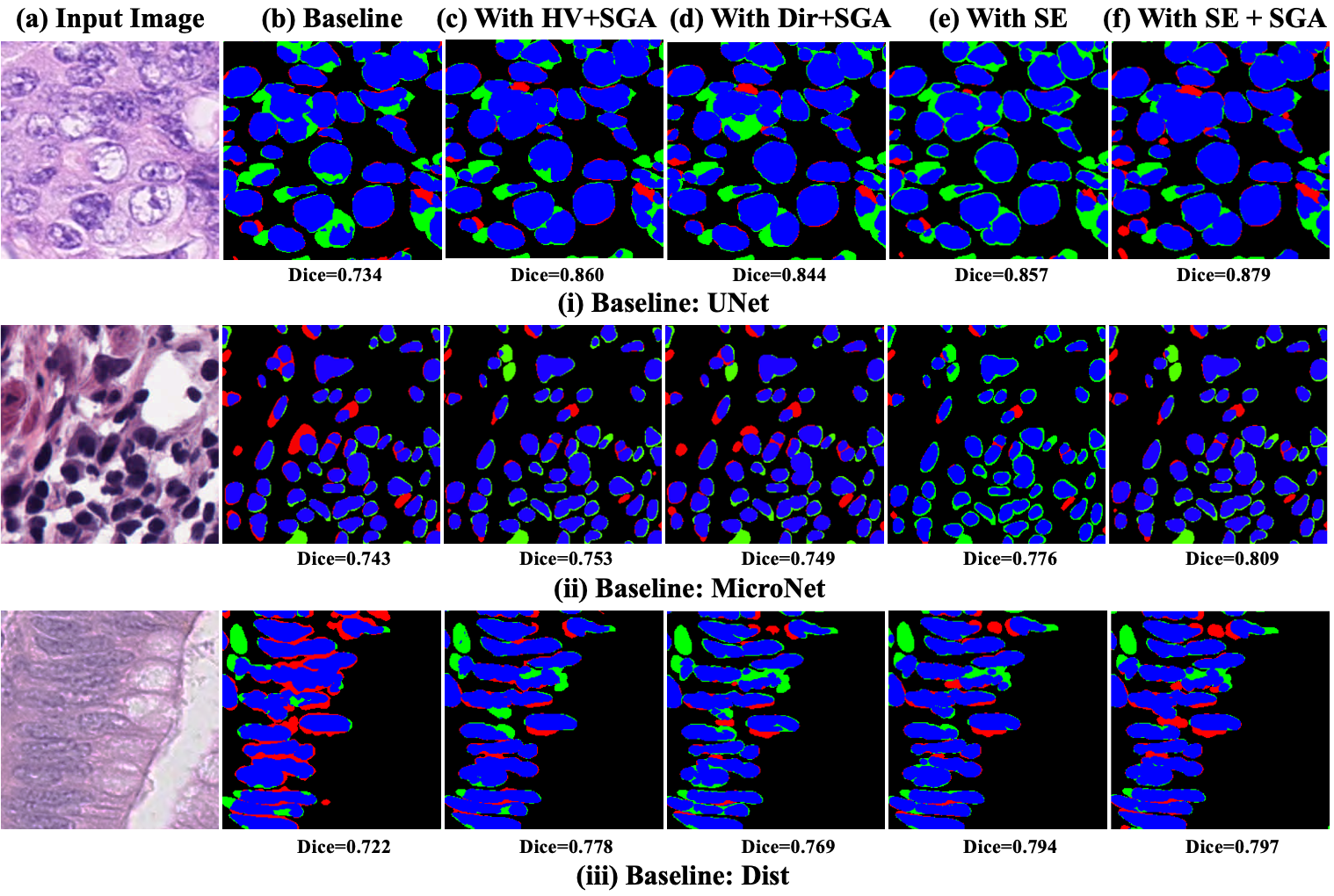}
	\caption{\textbf{The error analysis of different structure modeling methods based on UNet, MicroNet, and Dist.} ``Blue" is the true positive prediction. ``Red" is false positive error (FP) and ``Green" is the false negative error (FN).}
	\label{fig:fig8}
	\vspace{-0.6cm}
\end{figure}

\subsection{Ablation Studies}
To validate the effectiveness of each component employed in SEINE, we conduct a series of ablation experiments and analyses on the four datasets.
Throughout all experiments, the \textbf{best} result is bold, and the \underline{second best} is underlined.

\subsubsection{Effectiveness of contour-based SE}
To demonstrate the importance of structure modeling in nuclei segmentation, we employ UNet \cite{ronneberger2015u}, MicroNet \cite{raza2019micro}, and Dist \cite{naylor2018segmentation} as baselines and compare performances of different encoding strategies, including HV \cite{graham2019hover}, Dir \cite{he2021cdnet} and SE. The experimental results are shown in the first four rows of \textcolor{highlight}{\textbf{Table \ref{tab:ab1}}}. For fair comparisons, we add the contour prediction to the semantic branch, and the baseline performances are exhibited in the first row. From the second row to the fourth row, the lower-right values represent the increasing performance corresponding to baselines.

\begin{table}[t]
	\centering
	\caption{The ablation experiment results of SFF and PE modules on four different datasets (without SGA).}
	\renewcommand{\arraystretch}{1.15}
	\resizebox{\linewidth}{!}{
		\begin{tabular}{ccccccccc}
			\Xhline{1.4pt}
			\textbf{Datasets} & \textbf{Set.}& \textbf{SGA} & \textbf{SFF} & \textbf{PE} & \textbf{Dice}($\uparrow$) & \textbf{AJI}($\uparrow$) &  \textbf{HausD}($\downarrow$) &  \textbf{PQ}($\uparrow$) \\
			\Xhline{1.2pt}
			& \textbf{{\romannumeral1}}& \ding{56} & & & 0.764 & 0.491 & 12.74 &  0.441  \\
			& \textbf{{\romannumeral2}}& \ding{56} & \ding{52} &   &  0.809 & 0.529 & 10.61 & \underline{0.470}   \\
			\textbf{ConSeP} & \textbf{{\romannumeral3}} & \ding{56} &  & \ding{52} & \textbf{0.817} & \textbf{0.546} & \textbf{9.49} & \textbf{0.472}  \\
			& \textbf{{\romannumeral4}} &  \ding{56} & \ding{52} & \ding{52}& \underline{0.816} & \underline{0.542} & \underline{10.05} & \underline{0.470}  \\
			\Xhline{0.8pt}
			& \textbf{{\romannumeral1}}& \ding{56} & &   & 0.805 & 0.575 & 7.40 &0.525   \\
			& \textbf{{\romannumeral2}} & \ding{56} & \ding{52} & & 0.819 & 0.589 & \textbf{6.77} & \textbf{0.566}  \\
			\textbf{MoNuSeg} & \textbf{{\romannumeral3}} & \ding{56}& &\ding{52} & \underline{0.824} & \textbf{0.601} & 6.94 & 0.563  \\
			& \textbf{{\romannumeral4}} & \ding{56} & \ding{52} & \ding{52} & \textbf{0.829} & \underline{0.598} & \underline{6.78} & \underline{0.565} \\  
			\Xhline{0.8pt}
			& \textbf{{\romannumeral1}}& \ding{56} & & & 0.796 & 0.647 & 5.98&  0.587  \\
			& \textbf{{\romannumeral2}}& \ding{56} & \ding{52} &  &  0.824 & 0.666 & 5.60 & 0.607    \\
			\textbf{PanNuke} & \textbf{{\romannumeral3}} & \ding{56} & & \ding{52} & \textbf{0.827} & \textbf{0.672} & \textbf{5.14} & \textbf{0.612}  \\
			& \textbf{{\romannumeral4}} &  \ding{56} & \ding{52} & \ding{52} & \underline{0.826} & \underline{0.670} & \underline{5.51} & \underline{0.610}    \\
			\Xhline{0.8pt}
			& \textbf{{\romannumeral1}}& \ding{56} & &  & 0.765 & 0.612 & 6.96 & 0.597   \\
			& \textbf{{\romannumeral2}} & \ding{56} & \ding{52} &  & 0.794& 0.655 & 4.88 &0.615  \\
			\textbf{TNBC} & \textbf{{\romannumeral3}} & \ding{56}& &\ding{52} & \underline{0.799} & \underline{0.706} & \underline{4.06} & \underline{0.624}   \\
			& \textbf{{\romannumeral4}} & \ding{56} & \ding{52} & \ding{52} & \textbf{0.814} & \textbf{0.712} & \textbf{3.72} & \textbf{0.651} \\
			\Xhline{1.4pt}
	\end{tabular}}
	\label{tab:ab2}
\end{table}

From the results in \textcolor{highlight}{\textbf{Table \ref{tab:ab1}}}, we can obtain three points: (1) The use of HV encoding may result in a decline in performance, despite its superiority over Dir encoding. This decline is especially noticeable when applying the UNet model to the MoNuSeg dataset, as evidenced by a 2.54 increase in Hausdorff distance. (2) The Dir encoding has a varying impact on model performances. For instance, when adding Dir encoding to the UNet, the Dice improves by 1.2\% on the TNBC dataset. However, on ConSep and MoNuSeg datasets, the Dice decreases by 3.0\% and 6.7\%, respectively. (3) The models with SE achieve the best performance in comparison to HV and Dir encodings. In detail, the Dice demonstrates remarkable enhancements of 1.6\%, 0.8\%, 0.6\%, and 1.7\%, respectively across four datasets when using UNet. Similar improvements can also be observed in other models. Especially in Dist, the Dice on MonNuSeg obtains 2.8\% improvement. These results suggest that the use of HV and Dir may decrease performances of baselines due to conflicting feature representations between the semantic and structure branches.
In contrast, incorporating SE as the structure prediction task can significantly enhance the model's performance as it considers semantic consistency. 

\begin{figure}[t]
	\centering
	\includegraphics[width=0.965\linewidth]{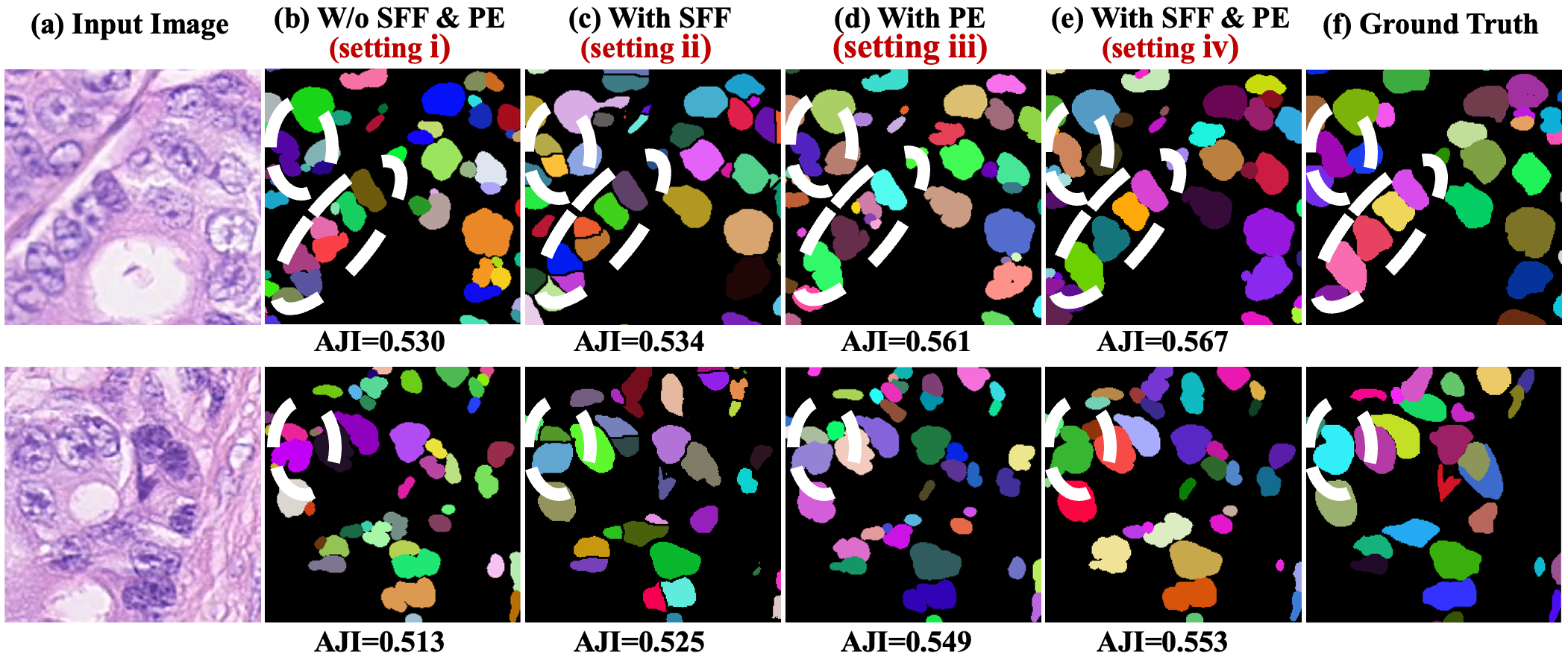}
	\caption{The qualitative analysis of SFF and PE modules.}%White circles indicate that the PE module reduces the fragmented instance segmentation and provides more accurate segmentation results. }
\label{fig:fig10}
\vspace{-0.5cm}
\end{figure}

\begin{table}[t]
	\centering
	\caption{The ablation experiment results of SFF and PE modules on four different datasets (with SGA).}
	\renewcommand{\arraystretch}{1.15}
	\resizebox{\linewidth}{!}{
		\begin{tabular}{ccccccccc}
			\Xhline{1.4pt}
			\textbf{Datasets} & \textbf{Set.}& \textbf{SGA} & \textbf{SFF} & \textbf{PE} & \textbf{Dice}($\uparrow$) & \textbf{AJI}($\uparrow$) &  \textbf{HausD}($\downarrow$) &  \textbf{PQ}($\uparrow$) \\
			\Xhline{1.2pt}
			& \textbf{{\romannumeral1}}& \ding{52} &  &  & 0.815& 0.536 & 9.21 &  0.465  \\
			& \textbf{{\romannumeral2}}& \ding{52} & \ding{52} &   &  \underline{0.832} & 0.551 & 8.81 & 0.492   \\
			\textbf{ConSeP} & \textbf{{\romannumeral3}} & \ding{52} &  & \ding{52} & 0.830 & \underline{0.564} & \textbf{8.66} & \underline{0.506}  \\
			& \textbf{{\romannumeral4}} &  \ding{52} & \ding{52} & \ding{52} & \textbf{0.834} & \textbf{0.566} & \underline{8.79}  &  \textbf{0.511}   \\
			\Xhline{0.8pt}
			& \textbf{{\romannumeral1}}& \ding{52} &  &   & 0.836 & 0.603 & 6.12 &0.574   \\
			& \textbf{{\romannumeral2}} & \ding{52} & \ding{52} & & 0.831& 0.608& 5.81 & \underline{0.589}  \\
			\textbf{MoNuSeg} & \textbf{{\romannumeral3}} & \ding{52}& &\ding{52} & \textbf{0.842} & \textbf{0.633} & \underline{5.03} &  0.587  \\
			& \textbf{{\romannumeral4}} & \ding{52} & \ding{52} & \ding{52} & \underline{0.840} &\underline{0.627} & \textbf{4.66} & \textbf{0.597} \\
			\Xhline{0.8pt}
			& \textbf{{\romannumeral1}}& \ding{52} &  &  & \textbf{0.827} & 0.670 & 4.86 &  0.611  \\
			& \textbf{{\romannumeral2}}& \ding{52} & \ding{52} &  &  \textbf{0.827} & 0.673 & \underline{4.48} & \textbf{0.616}    \\
			\textbf{PanNuke} & \textbf{{\romannumeral3}} & \ding{52} & & \ding{52} & \underline{0.822} & \textbf{0.682} &  4.52& \underline{0.615}   \\
			& \textbf{{\romannumeral4}} &  \ding{52} & \ding{52} & \ding{52}& \textbf{0.827} & \underline{0.680} & \textbf{4.44} & \textbf{0.616}   \\
			\Xhline{0.8pt}
			& \textbf{{\romannumeral1}}& \ding{52} &  &  & 0.789 & 0.679 & 4.98 & 0.636   \\
			& \textbf{{\romannumeral2}} & \ding{52} & \ding{52} & & \underline{0.816} & 0.701& \underline{3.71} & \textbf{0.645}  \\
			\textbf{TNBC} & \textbf{{\romannumeral3}} & \ding{52}& &\ding{52} & 0.806 & \textbf{0.717} & 3.81 & 0.622  \\
			& \textbf{{\romannumeral4}} & \ding{52} & \ding{52} & \ding{52} & \textbf{0.818} & \underline{0.715} & \textbf{3.20} & \underline{0.643} \\
			\Xhline{1.4pt}
	\end{tabular}}
	\label{tab:ab3}
\end{table}

\begin{figure}[t]
	\centering
	\includegraphics[width=0.98\linewidth]{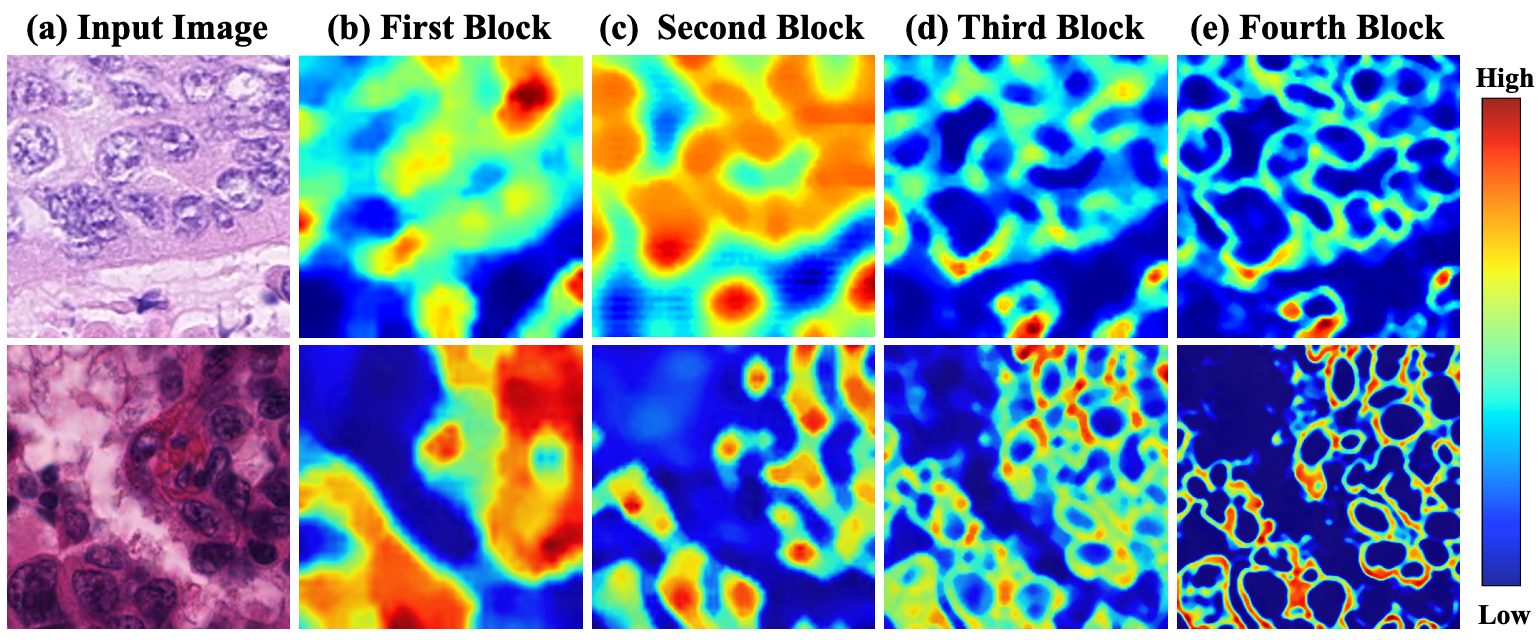}
	\caption{The global attention maps generated by different blocks.}
	\label{fig:fig11}
	\vspace{-0.5cm}
\end{figure}

\begin{figure*}[t]
	\centering
	\includegraphics[width=0.98\linewidth]{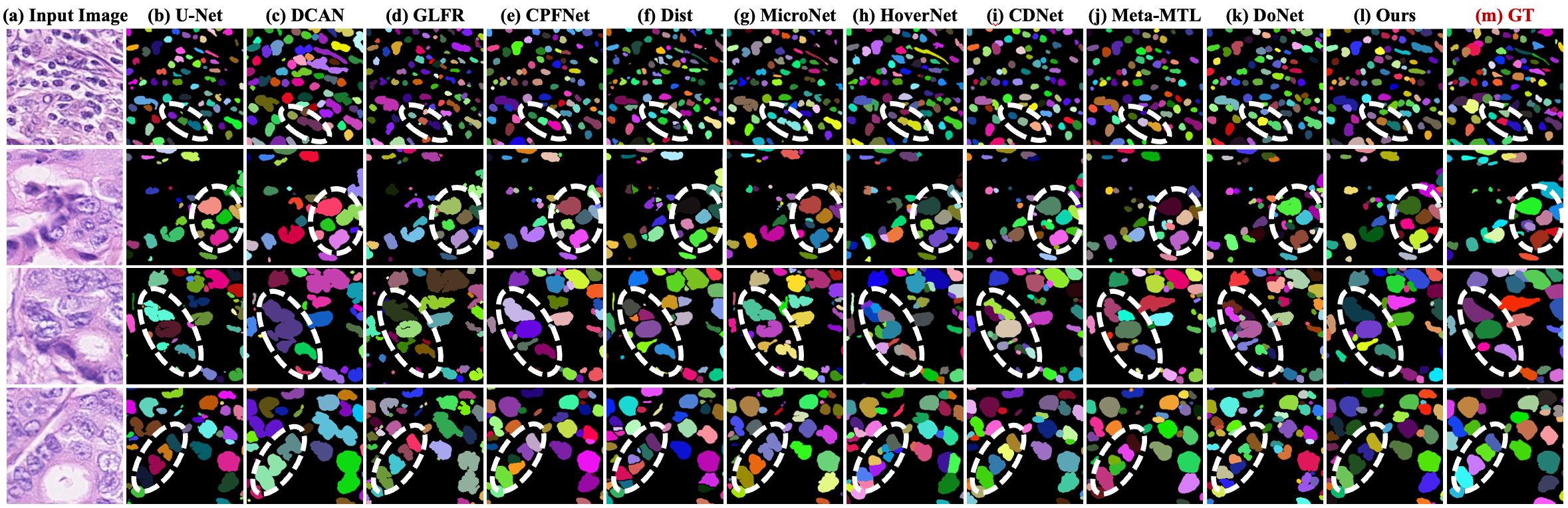}
	\caption{\textbf{The qualitative comparision results of various methods.} 
		Different colors indicate different instances in the images.  White circles are drawn for clear comparisons.}
	\label{fig:fig7}
	\vspace{-0.6cm}
\end{figure*}

\begin{table}[t]
	\centering
	\caption{\textbf{The hyperparameter analysis of numbers of SGA modules.} \textit{No.L} is the abbreviation of number of layers. $L=0$ represents our baseline.}
	\label{tab:ab6}
	\renewcommand{\arraystretch}{1.2}
	\resizebox{\linewidth}{!}{
		\begin{tabular}{c|cccc|cccc}
			\Xhline{1.4pt}
			\multirow{2}{*}{\textit{\textbf{No.L}}} & \multicolumn{4}{c}{\textbf{ConSeP}} & \multicolumn{4}{c}{\textbf{MoNuSeg}}\\
			\cline{2-9}
			& \textbf{Dice($\uparrow$)} & \textbf{AJI($\uparrow$)} &  \textbf{HausD($\downarrow$)}&  \textbf{PQ($\uparrow$)} & \textbf{Dice($\uparrow$)} & \textbf{AJI($\uparrow$)} &  \textbf{HausD($\downarrow$)} &  \textbf{PQ($\uparrow$)}\\
			\Xhline{1.0pt}
			\textbf{0} &  0.816 & 0.542 & 10.05 & 0.470 &  0.829 & 0.598 & 6.78 & 0.565 \\
			\textbf{1} & 0.821& 0.542&10.20 & 0.489 & 0.831 & 0.607 & 6.44 & 0.573 \\
			\textbf{2} &  \textbf{0.834} & \textbf{0.566} & \textbf{8.79}  & \textbf{0.511} & \underline{0.840} & \textbf{0.627} & \underline{4.66}  & \textbf{0.597} \\
			\textbf{3} & \underline{0.832}& \underline{0.548}& \underline{8.89} & \underline{0.498} & \textbf{0.841} & \underline{0.625} & \textbf{4.61} & \underline{0.587} \\
			\Xhline{1.0pt}
			\multirow{2}{*}{\textit{\textbf{No.L}}} & \multicolumn{4}{c}{\textbf{PanNuke}} & \multicolumn{4}{c}{\textbf{TNBC}}\\
			\cline{2-9}
			& \textbf{Dice($\uparrow$)} & \textbf{AJI($\uparrow$)} &  \textbf{HausD($\downarrow$)}&  \textbf{PQ($\uparrow$)} & \textbf{Dice($\uparrow$)} & \textbf{AJI($\uparrow$)} &  \textbf{HausD($\downarrow$)} &  \textbf{PQ($\uparrow$)}\\
			\Xhline{1.0pt}
			\textbf{0} & 0.816 & 0.664 & 5.89 & 0.604 & 0.804 & 0.692 & 4.20 & 0.632\\
			\textbf{1} & 0.825 & \underline{0.673} & 5.48 & \underline{0.614} & 0.810 & 0.706 & 3.98 & \textbf{0.644} \\
			\textbf{2} & \textbf{0.827} & \textbf{0.680} & \textbf{4.44} & \textbf{0.616} & \textbf{0.818} & \textbf{0.715} & \textbf{3.20}  & \underline{0.643} \\
			\textbf{3} & \underline{0.826}& 0.671 & \underline{4.96} & 0.610 & \underline{0.817} & \underline{0.711} & \underline{3.42} & 0.639 \\
			\Xhline{1.4pt}
	\end{tabular}}
	\vspace{-0.2cm}
\end{table}

\subsubsection{Effectiveness of SGA module}
To validate the impact of the SGA, we compare the experimental results before and after adding the SGA module to UNet+SE, MicroNet+SE, and Dist+SE models as listed in the last row of \textbf{\textcolor{highlight}{Table \ref{tab:ab1}}}. The lower-right values indicate the performance improvements achieved when incorporating the SGA module. 
Comparing the performance between adding SE and SE+SGA shows that the SGA module significantly improves the model's performance. 
On the PanNuke dataset, the UNet model exhibits notable improvements in Dice, AJI, and PQ scores, with increases of 2.3\%, 4.9\%, and 4.4\%, respectively. Additionally, the HausD decreases by 2.25. Similar conclusions can be drawn on other datasets and baselines, demonstrating the effectiveness of our SGA in guiding semantic feature learning. However, when using HV or Dir, the model performance may decrease.

To analyze the reasons behind the performance variations, we plot the attention maps of the foreground pixel (\textcolor{green}{+}) and background pixel (\textcolor{blue}{×}) shown in \textcolor{highlight}{\textbf{Fig. \ref{fig:fig9}}}. 
From the attention maps, we can see: (1) In HV encoding, the foreground pixel tends to pay attention to the left area. In Dir encoding, the foreground focuses on the part regions of nuclei. But in our SE, the foreground pixel focuses on the contour and foreground regions. This result is consistent with the structural modeling characteristics. HV uses horizontal distance modeling, so the foreground pixel has a higher correlation with those areas that are the same distance to the centroid. Dir uses a directional category structure modeling method, so it can only focus on areas with the same category. In contrast, our method focuses on regions that are at the same distance to the contour and, therefore, are highly correlated with contour pixels. This shows that our method considers structural similarity more comprehensively. (2) In HV and Dir encodings, the background pixel pays attention to some irrelevant regions. But our SE still focuses on nuclei contour. These observations demonstrate HV and cannot learn complete structure information. However, our SE reasonably models the structure of nuclei.

\textcolor{highlight}{\textbf{Fig. \ref{fig:fig8}}} shows the error analysis visualization. From the figure we can see the model with SE+SGA has lower FP errors (red) and higher Dice values than other methods, which further indicates the SE exactly encodes the nuclei structure and the SGA helps structure interaction between nuclei, which strengthens the integrality of nuclei. Therefore, our method can effectively solve the under-segmentation problem.

%It is worth noting that the performance after adding SGA depends on the choice of structure modeling methods. Only when the nuclei structure is encoded properly can the SGA enhance the performance of the model. The specific analysis of this aspect will be discussed in the discussion section. 

\subsubsection{Effectiveness of SFF}
To verify the effectiveness of the SFF module, we conduct ablation experiments as shown in \textcolor{highlight}{\textbf{Table \ref{tab:ab2}}} and \textcolor{highlight}{\textbf{\ref{tab:ab3}}}. The former is the ablation results with SGA, and the latter is without SGA.
(1) Comparing the settings \textbf{\romannumeral1} and \textbf{\romannumeral2} without SGA, we can observe a significant improvement after using SFF,  with the Dice increasing by 4.5\%, 1.4\%, 2.8\%, and 2.9\% on the four datasets, respectively.
(2) When using SGA, the performances still have an improvement. For instance, on ConSeP and TNBC datasets, the AJI increases by 1.5\% and 2.2\%, respectively.  
These observations indicate that in the absence of SGA, the SFF can establish correlations between the semantic and the structure branches. When SGA is introduced, SFF continues to enhance the semantic representation capability of the structure branch, thereby promoting model performance simultaneously.
We also conduct a visualization experiment for the SFF shown in \textcolor{highlight}{\textbf{Fig. \ref{fig:fig10}}}. From the figure, the model without SFF always exhibits coarse boundaries and seriously fragmented segmentation. When adding SFF, the problems are eased because the SFF enhances the consistency between structure and semantic branches.

\subsubsection{Effectiveness of PE}
To verify the effectiveness of the PE module, we conduct ablation experiments shown in settings \textbf{\romannumeral3} and \textbf{\romannumeral4} of \textcolor{highlight}{\textbf{Table \ref{tab:ab2}}}. From the results, on ConSeP, MoNuSeg, and PanNuke datasets, the AJI has the best performances corresponding to 0.546, 0.601 and 0.672 when using PE. 
Meanwhile, the PE module visualization results are shown in \textcolor{highlight}{\textbf{Fig. \ref{fig:fig10}}}. In the figure, the incorrect contour prediction is rectified when the PE module is added. That is because when without the PE, the model tends to mistake some darker textures as contours, resulting in a shift of nuclei centroid in fragmented prediction. However, when the PE is employed, the model actively learns the nuclei centroid, suppressing fragmented segmentation. This indicates that PE enhances the integrality of nuclei using the effective contour correction method.

\begin{table}[t]
	\centering
	\caption{\textbf{The hyperparameter analysis of the layer number and the block number of SGA.} ``i-th Blo." represents that we use SGA from the i-th to the 4-th block.}%``Blo" is the abbreviation of Block. 
\label{tab:ab7}
\renewcommand{\arraystretch}{1.3}
\resizebox{\linewidth}{!}{
	\begin{tabular}{c|cccc|cccc}
		\Xhline{1.4pt}
		\multirow{2}{*}{\textbf{\textit{No.L}}}&  \multicolumn{4}{c|}{\textbf{Dice($\uparrow$)}} & 
		\multicolumn{4}{c}{\textbf{AJI($\uparrow$)}}  \\ 
		\cline{2-9}
		& \textbf{1st Blo.} & \textbf{2nd Blo.} & \textbf{3rd Blo.} & \textbf{4th Blo.} & \textbf{1st Blo.} & \textbf{2nd Blo.} & \textbf{3rd Blo.} & \textbf{4th Blo.} \\
		\Xhline{1.0pt}
		\textbf{0} & 0.808 & 0.816 & 0.804 & 0.798 & 0.517 & 0.542 & 0.520 & 0.518 \\
		\textbf{1} & 0.810 & 0.821 &0.812 & 0.806 & 0.529 & 0.542 & \underline{0.538} & 0.527   \\
		\textbf{2} & \textbf{0.818} & \underline{0.834} & \textbf{0.819} & \textbf{0.811} & \textbf{0.536} & \textbf{0.566} & \textbf{0.549} & \underline{0.540} \\
		\textbf{3} & \underline{0.817}& \textbf{0.835} & \underline{0.817} & \underline{0.810}  & \underline{0.532} & \underline{0.548} & 0.537 & \textbf{0.541} \\
		\Xhline{1.0pt}
		\multirow{2}{*}{\textbf{\textit{No.L}}} &  \multicolumn{4}{c|}{\textbf{HausD($\downarrow$)}} & 
		\multicolumn{4}{c}{\textbf{PQ($\uparrow$)}}  \\ 
		\cline{2-9}
		& \textbf{1st Blo.} & \textbf{2nd Blo.} & \textbf{3rd Blo.} & \textbf{4th Blo.} & \textbf{1st Blo.} & \textbf{2nd Blo.} & \textbf{3rd Blo.} & \textbf{4th Blo.} \\
		\Xhline{1.0pt}
		\textbf{0} &  12.49 & 10.05 & 11.89 & 11.54 & 0.434 & 0.470 & 0.450 & 0.443\\
		\textbf{1} & 12.07 & 10.20 & \underline{11.20} & 11.47 & 0.441 & 0.489 & 0.446 & 0.438 \\
		\textbf{2} & \underline{11.34} & \textbf{8.79} & \textbf{10.97} & \textbf{10.8}  & \textbf{0.457} & \textbf{0.511} &  \textbf{0.507} & \textbf{0.496} \\
		\textbf{3} & \textbf{9.77} & \underline{8.89} & 11.49 & \underline{11.20} & \underline{0.454} & \underline{0.498} & \underline{0.499} & \underline{0.480}\\
		\Xhline{1.4pt}
\end{tabular}}
\vspace{-0.2cm}
\end{table}

\begin{table*}[t]
	\caption{The quantitative comaparison of different methods on four datasets. The numerical numbers are mean values and the corresponding subscript is the standard deviation of cross-validation calculation on test set. ``$\dag$" represents p-value of AJI \textless 0.001. ``*" represents p-value of AJI  \textless 0.05.}\label{tab:comp}
	\vspace{-0.5cm}
		\begin{center}
		\renewcommand{\arraystretch}{1.43}
		\resizebox{\linewidth}{!}{
			\begin{tabular}{c|cccc|cccc|cccc|cccc}
				\Xhline{1.4pt}
				\multirow{2}{*}{\textbf{Methods}}  &  \multicolumn{4}{c|}{\textbf{ConSeP}} & \multicolumn{4}{c|}{\textbf{MoNuSeg}}& \multicolumn{4}{c|}{\textbf{PanNuke}} & \multicolumn{4}{c}{\textbf{TNBC}} 
				\\ \cline{2-17}
				& \textbf{Dice(\%) $\uparrow$} &\textbf{AJI(\%)$\uparrow$} & \textbf{HausD$\downarrow$} & \textbf{PQ(\%)$\uparrow$} & \textbf{Dice(\%)$\uparrow$} & \textbf{AJI(\%)$\uparrow$} & \textbf{HausD$\downarrow$} & \textbf{PQ(\%)$\uparrow$}& \textbf{Dice(\%)$\uparrow$ }& \textbf{AJI(\%)$\uparrow$} & \textbf{HausD$\downarrow$} & \textbf{PQ(\%)$\uparrow$} & \textbf{Dice(\%)$\uparrow$} & \textbf{AJI(\%)$\uparrow$} & \textbf{HausD$\downarrow$} & \textbf{PQ(\%)$\uparrow$}  \\ 
				\Xhline{1.2pt}
				\textbf{UNet}$^{*}$ \cite{ronneberger2015u} &76.4$\pm$\scriptsize2.1 & 49.1$\pm$\scriptsize1.6 & 15.90 $\pm$\scriptsize 3.21 & 44.1$\pm$\scriptsize1.8 & 82.5 $\pm$\scriptsize1.8 & 57.5$\pm$\scriptsize 2.3 & 6.66$\pm$\scriptsize2.71 & 52.5$\pm$\scriptsize2.0 & 76.2$\pm$\scriptsize1.6 & 58.7$\pm$\scriptsize1.8 & 8.74$\pm$\scriptsize2.94 & 56.7$\pm$\scriptsize2.5 & 79.6$\pm$\scriptsize3.2 & 65.4$\pm$\scriptsize1.5 & 7.89$\pm$\scriptsize1.71 & 60.5$\pm$\scriptsize2.3 \\
				\textbf{DCAN}$^{*}$ \cite{chen2016dcan} & 78.7$\pm$\scriptsize2.5 & 36.0$\pm$\scriptsize2.0 & 18.83$\pm$\scriptsize4.14 & 32.1$\pm$\scriptsize1.8 & 81.2$\pm$\scriptsize3.5 & 46.4$\pm$\scriptsize4.6 & 11.21$\pm$\scriptsize2.11 & 42.7$\pm$\scriptsize2.3 & 74.0$\pm$\scriptsize2.6 & 50.4$\pm$\scriptsize3.1 & 11.61$\pm$\scriptsize1.22 & 46.0$\pm$\scriptsize5.8 & 70.4$\pm$\scriptsize4.1 & 49.7$\pm$\scriptsize4.4 & 10.87$\pm$\scriptsize2.32 & 43.3$\pm$\scriptsize5.5   \\
				$\textbf{GLFR}^{*}$ \cite{song2022global}  & 77.4$\pm$\scriptsize4.6 & 40.1$\pm$\scriptsize3.2 & 16.30$\pm$\scriptsize4.50 & 34.6$\pm$\scriptsize3.8 & 80.6$\pm$\scriptsize2.9 & 48.9$\pm$\scriptsize3.8 & 10.33$\pm$\scriptsize3.20 & 44.6$\pm$\scriptsize1.8 & 76.7$\pm$\scriptsize2.1 & 54.3$\pm$\scriptsize2.6 & 10.76$\pm$\scriptsize2.40 & 52.4$\pm$\scriptsize1.9 & 75.3$\pm$\scriptsize3.3 & 60.1$\pm$\scriptsize2.7 & 9.76$\pm$\scriptsize3.43 & 54.4$\pm$\scriptsize4.1 \\
				\textbf{CPFNet}$^{\dag}$ \cite{feng2020cpfnet} & 81.8$\pm$\scriptsize3.2 & 53.5$\pm$\scriptsize2.9 & 12.87$\pm$\scriptsize3.30 & 48.1$\pm$\scriptsize2.8 & 80.2$\pm$\scriptsize4.3 & 56.5$\pm$\scriptsize2.4 & 8.78$\pm$\scriptsize4.60 & 49.4$\pm$\scriptsize2.0 & 80.8$\pm$\scriptsize2.4 & 62.2$\pm$\scriptsize3.9 & 9.44$\pm$\scriptsize3.13 & 58.4$\pm$\scriptsize1.8 & 79.1$\pm$\scriptsize3.0 & 66.3$\pm$\scriptsize3.1 & 5.90$\pm$\scriptsize4.67 & 62.1$\pm$\scriptsize2.2 \\
				\textbf{Dist}$^{\dag}$ \cite{naylor2018segmentation}  & 82.7$\pm$\scriptsize1.6 & 51.8$\pm$\scriptsize2.2 & 14.98$\pm$\scriptsize3.21 & 42.2$\pm$\scriptsize2.0 & 83.1$\pm$\scriptsize1.7 & 57.2$\pm$\scriptsize1.1 & 6.89$\pm$\scriptsize2.33 & 50.0$\pm$\scriptsize3.2 & \textbf{83.0$\pm$\scriptsize1.9} & 64.8$\pm$\scriptsize1.8 & 8.07$\pm$\scriptsize2.89 & 55.2$\pm$\scriptsize1.9 & 78.6$\pm$\scriptsize4.1 & \underline{68.1$\pm$\scriptsize2.4} & 6.01$\pm$\scriptsize3.57 & 61.1$\pm$\scriptsize1.7 \\
				\textbf{MicroNet}$^{*}$ \cite{raza2019micro}  & 80.5$\pm$\scriptsize1.3 & 51.7$\pm$\scriptsize1.6 & 13.88$\pm$\scriptsize3.06 & 45.7$\pm$\scriptsize1.3& 81.8$\pm$\scriptsize3.4 & 50.2$\pm$\scriptsize1.6 & 9.44$\pm$\scriptsize1.37 & 46.7$\pm$\scriptsize1.8 & 81.3$\pm$\scriptsize2.0 & 63.5$\pm$\scriptsize2.8 & 9.57$\pm$\scriptsize2.63 & 56.4$\pm$\scriptsize1.7 & 81.3$\pm$\scriptsize2.9 & 67.3$\pm$\scriptsize3.2 & 7.09$\pm$\scriptsize3.28 & 60.5$\pm$\scriptsize1.5 \\
				\textbf{HoverNet}$^{\dag}$ \cite{graham2019hover}  & \textbf{83.7$\pm$\scriptsize1.4} & 
				51.3$\pm$\scriptsize1.7 & 10.26$\pm$\scriptsize2.88 & 48.2$\pm$\scriptsize1.4& \underline{83.2$\pm$\scriptsize3.1} &\underline{59.2$\pm$\scriptsize1.6} & 6.96$\pm$\scriptsize1.01 & 55.7$\pm$\scriptsize1.9 & 82.4$\pm$\scriptsize1.5 & 65.7$\pm$\scriptsize2.1 & 7.81$\pm$\scriptsize1.88 & 60.1$\pm$\scriptsize0.9 & \underline{81.5$\pm$\scriptsize2.8} & 66.2$\pm$\scriptsize2.9 & 6.93$\pm$\scriptsize3.11 & 62.4$\pm$\scriptsize2.0 \\
				\textbf{CDNet}$^{\dag}$ \cite{he2021cdnet} & 
				82.2$\pm$\scriptsize2.1&
				\underline{54.8$\pm$\scriptsize2.2} & \underline{9.17$\pm$\scriptsize3.07} & \underline{48.8$\pm$\scriptsize1.4} & 
				82.0$\pm$\scriptsize2.1 & 
				58.4$\pm$\scriptsize0.9  & 
				\underline{5.17$\pm$\scriptsize1.13} & 
				55.5$\pm$\scriptsize1.1 & 
				82.2$\pm$\scriptsize1.4 & 
				\underline{67.7$\pm$\scriptsize1.8} & \underline{4.91$\pm$\scriptsize1.90} & 
				\textbf{61.9$\pm$\scriptsize1.2} & 
				\textbf{81.8$\pm$\scriptsize2.4} & 
				72.4$\pm$\scriptsize2.6 & 
				\underline{4.11$\pm$\scriptsize2.87} & \underline{63.5$\pm$\scriptsize2.2}  \\
				\textbf{Meta-MTL}$^{\dag}$\cite{han2022meta}  & 81.3$\pm$\scriptsize2.6 & 50.5$\pm$\scriptsize3.0 & 10.06$\pm$\scriptsize1.17 & 46.0$\pm$\scriptsize2.1 & 81.7$\pm$\scriptsize1.9 & 57.4$\pm$\scriptsize2.1 & 6.89$\pm$\scriptsize2.31 & 52.9$\pm$\scriptsize1.4 & 81.3$\pm$\scriptsize1.8 & 67.4$\pm$\scriptsize1.2 & 5.12$\pm$\scriptsize2.43& 59.8$\pm$\scriptsize2.1 & 78.5$\pm$\scriptsize2.9 & 65.3$\pm$\scriptsize2.6 & 7.34$\pm$\scriptsize2.73 & 63.3$\pm$\scriptsize1.9 \\
				\textbf{DoNet}$^{*}$ \cite{jiang2023donet} & 80.8$\pm$\scriptsize2.2 & 52.4$\pm$\scriptsize2.8 & 11.71$\pm$\scriptsize2.94 & 45.2$\pm$\scriptsize2.3 & 81.2$\pm$\scriptsize2.1 & 57.1$\pm$\scriptsize1.9 & 7.98$\pm$\scriptsize2.24 & \underline{56.0$\pm$\scriptsize2.4} & 82.3$\pm$\scriptsize1.4 & 66.5$\pm$\scriptsize2.0 & 5.84$\pm$\scriptsize2.81 & 61.1$\pm$\scriptsize1.2 & 79.2$\pm$\scriptsize2.4 & 67.7$\pm$\scriptsize1.8 & 5.06$\pm$\scriptsize2.11 & 63.4$\pm$\scriptsize1.8 \\
				\hline
				\textbf{SEINE}$^{\dag}$ &  \underline{83.4$\pm$\scriptsize1.7} & \textbf{56.6{$\pm$\scriptsize2.4}} & \textbf{8.79{$\pm$\scriptsize3.15}} &  \textbf{51.1{$\pm$\scriptsize1.6}} & \textbf{84.0{$\pm$\scriptsize1.8}} & \textbf{62.7{$\pm$\scriptsize1.3}} & \textbf{4.66{$\pm$\scriptsize2.05}} & \textbf{59.7{$\pm$\scriptsize0.9}} & \underline{82.7$\pm$\scriptsize1.3} & \textbf{68.0}$\tiny{\pm1.2}$ & \textbf{4.44{$\pm$\scriptsize2.18}} & \underline{61.6}$\tiny{\pm1.4}$ & \textbf{81.8{$\pm$\scriptsize2.2}} & \textbf{71.5{$\pm$\scriptsize1.9}} & \textbf{3.20{$\pm$\scriptsize2.61}} & \textbf{64.3{$\pm$\scriptsize2.2}}\\
				\Xhline{1.4pt}
		\end{tabular}}
	\end{center}
	\vspace{-0.4cm}
\end{table*}

\subsection{Hyperparameters Analysis}

We first investigate the impact of the layer number of SGA as shown in \textcolor{highlight}{\textbf{Table \ref{tab:ab6}}}. 
From the table, we can find that the introducing of SGA improves the segmentation results on four datasets, e.g., the Dice increases from 0.816 to 0.821 on ConSeP. These improvements indicate SGA is a useful module for promoting instance segmentation. Besides, the model performs best when $L=2$. Hence, we set $L=2$ as the default in all experiments. 

To verify the joint impact of the number of layers $L$ and the blocks involved in SGA, we conducted experiments with parameter combinations on ConSeP, and the results are shown in \textcolor{highlight}{\textbf{Table \ref{tab:ab7}}}. 
In the table, each row represents the layer number of SGA, ranging from 0 to 3; each column represents the first block inserted by SGA.
Based on the results, we can observe that when $L=2$ and blocks are set from $2$ to $4$, AJI, HausD, and PQ consistently reach the best performance. 
This is because when the first block is used for structure interaction, the resolution of the feature map is too low to depict nuclei position and contour, posing difficulties for network training. 

To intuitively illustrate the point, we draw global attention maps from the first block to the fourth block in \textcolor{highlight}{\textbf{Figure \ref{fig:fig11}}}. We can observe that the first and the second blocks focus on foreground features; while the third and the fourth blocks emphasize nuclei contours, as deeper blocks could capture more detailed information. Furthermore, the attention regions are coarse in column (b), which indicates the feature map of a small scale may bring difficulties for structure interactions.

\subsection{Comparison with the State-of-the-Art Methods}

We compare our method with ten SOTA methods, including UNet \cite{ronneberger2015u}, DCAN \cite{chen2016dcan}, GLFR \cite{song2022global}, CPFNet \cite{feng2020cpfnet}, Dist \cite{naylor2018segmentation}, MicroNet \cite{raza2019micro}, HoverNet \cite{graham2019hover}, CDNet \cite{he2021cdnet}, Meta-MTL \cite{han2022meta} and DoNet \cite{jiang2023donet}. 
Among these methods, HoverNet and CDNet model the nuclei structure based on the relative position to the centroid. Dist models the nuclei structure based on the contour, but it does not consider the structure interaction between semantics and structure. From the comparison results illustrated in \textcolor{highlight}{\textbf{Table \ref{tab:comp}}}, we can find that (1) SEINE outperforms other methods and achieves the SOTA performance on all datasets. On the ConSeP and TNBC datasets, our method obtains improvements in AJI, DQ, and PQ by at least 1\% compared to other methods. This is because our structure modeling strategy takes the location relationship of intranuclear and extranuclear areas into consideration and unifies the structure and semantic features. (2) On the MoNuSeg dataset, our method derives huge improvements in terms of AJI, DQ, and PQ, which surpass the second-best method by about 3\%. This is due to MoNuSeg dataset containing nuclei of diverse types, and our method could effectively capture the nuclei morphology by structure modeling. (3) The Dist, HoverNet, and CDNet achieve suboptimal or even SOTA performance on some datasets, further indicating the necessity of structure modeling and structure interaction. 

To qualitatively evaluate the performance of our method, we present a visualization comparison in \textcolor{highlight}{\textbf{Fig. \ref{fig:fig7}}}. From the visual results, we can find that existing nuclei segmentation methods cannot handle situations, such as the texture of nuclei being similar to that of surrounding tissue. These lead to instances of nuclei that can not be separated (e.g., DCAN) or issues with blurry boundaries (e.g., UNet, GLFR, and MicroNet). Even though some methods perform well in segmenting challenging chromophobe nuclei, they still exhibit deficiencies when faced with overlapping nuclei (e.g., HoverNet and CDNet). In contrast, our method derives excellent segmentation results under various scenes. These improvements can be attributed to the incorporation of structure modeling and structure interaction, which lead to more reasonable nuclei structures.

\begin{figure}[t]
	\centering
	\includegraphics[width=1.0\linewidth]{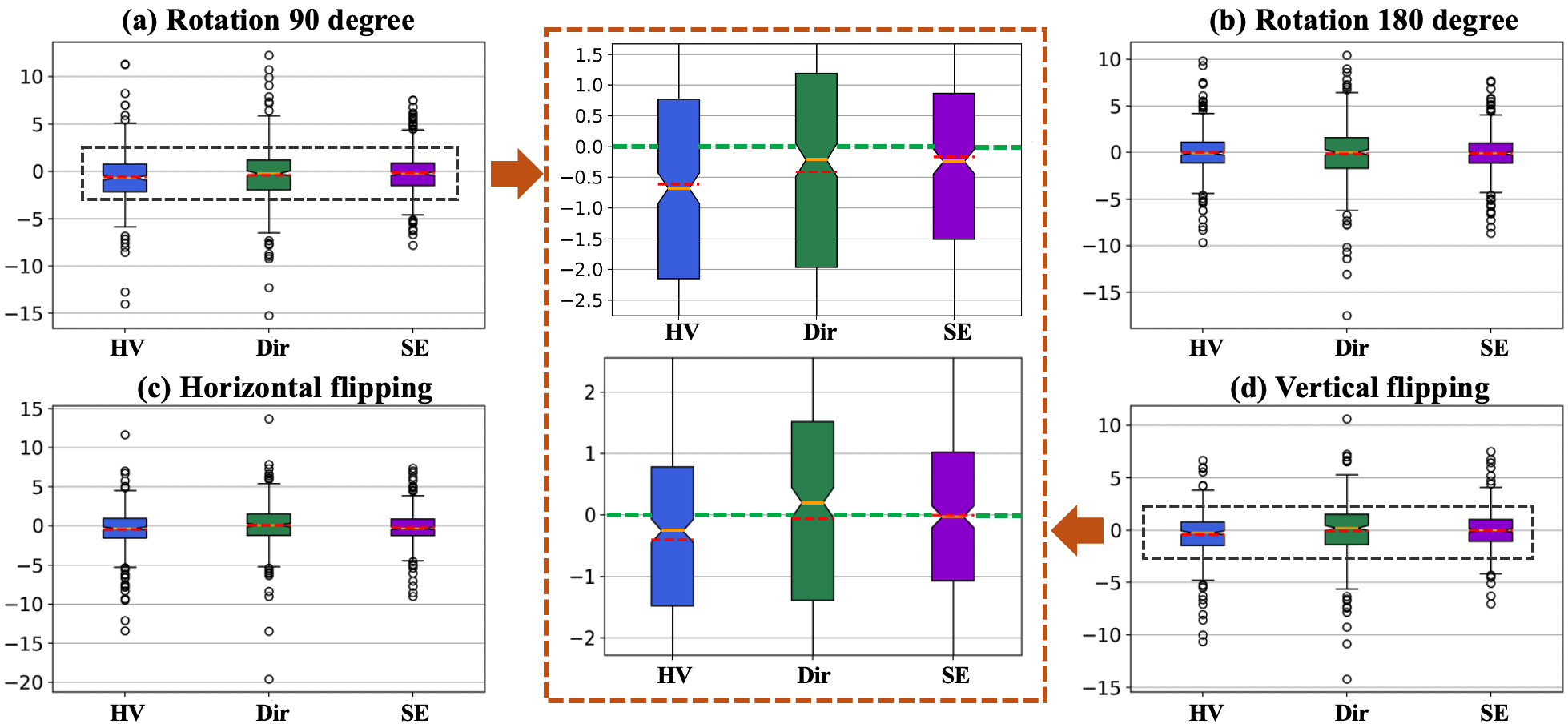}
	\caption{\textbf{The Dice bias caused by direction changes for different structure modeling methods. }  The yellow line in box represents median value and the red dotted line in box represents mean value.}
	\label{fig:fig14}
	\vspace{-0.4cm}
\end{figure}

%\section{Disussion and Conclusion}\label{conclu}

\section{Dicussion}\label{discuss}
In this part, we discuss the importance of direction-invariant structure modeling and the relations of different structure encoding methods. For the first problem, we conduct experiments on ConSeP and present the bias distributions in \textcolor{highlight}{\textbf{Fig. \ref{fig:fig14}}}. The box plots represent the Dice differences between image $I$ and the augmented image $Aug(I)$. The differences in the plots are calculated by: 
\vspace{-0.15cm}
\begin{equation}
	Bias_{Dice} = Net(Aug(I)) - Net(I),
%	\vspace{-0.2cm}
\end{equation}
where $Net(\cdot)$ represents a segmentation network.

Four data augmentation methods are employed in the exploratory experiments: 90-degree rotation, 180-degree rotation, horizontal flipping, and vertical flipping. From the figures, we can observe that HV and Dir encodings exhibit higher biases and variances. This phenomenon could be attributed to the lack of direction-invariant in HV and Dir, resulting in high sensitivity to changes in nuclei structure. The segmentation results vary a lot, even in different views of the same patch. In contrast, the biases of our SE are closer to zero and exhibit lower variance. This indicates that our SE is robust for direction changes.

Essentially, our SE is a generalized encoding method from which other encoding techniques could be deduced through SE. For instance, the HV and Dir encodings could be performed through the following formulations:
\vspace{-0.1cm}
\begin{equation}
	\begin{split}
		& HV_{h} \approx \nabla_h (y_{str}^{i,j}),\  HV_{v} \approx \nabla_v (y_{str}^{i,j}), \\
		& Dir \approx arctan(y_{str}^{i,j}),
	\end{split}
\end{equation}
where $HV_{h}$ and $HV_{v}$ represent the horizontal direction encoding and the vertical direction encoding of HV, respectively. $\nabla_h$ and $\nabla_v$ denote the horizontal and vertical differentiations. 

%\begin{figure}[t]
%	\centering
%	\includegraphics[width=0.85\linewidth]{Fig15.png}
%	\caption{\textbf{The failure cases of our proposed SINE.} There are serious under-segmentation when facing to fiber cells.}
%	\label{fig:fig15}
%	\vspace{-0.4cm}
%\end{figure}

%\subsection{The Limitations in SEINE}
%To analyze the limitations of our method, we presente some segmentation failure cases as shown in \textcolor{highlight}{\textbf{Fig. \ref{fig:fig15}}}. From the input images, we can see the fiber cells exhibit elongated shapes.Moreover, there are no other types of nuclei present simultaneously to assist in the structural learning of the fiber cells. As a result, the network can only focus on some foreground regions with similar textures and leads to significant segmentation deficiencies. In the future, staining separation may be a viable solution for addressing the issue of lightly stained fiber cells.

\section{Conclusion}\label{conclu}
This paper proposes a novel structure encoding and interaction network (SEINE) to overcome the challenge of under-segmentation in nuclei instance segmentation, especially chromophobe nuclei. Specifically, SEINE introduces a new contour-based structure encoding scheme considering the correlation between structure and semantics. With the encoding, structure-guided attention is applied to enhance structure learning, taking the clear nuclei as prototypes. To further enhance the feature consistency between the semantic branch and the structure branch, we propose a semantic feature fusion method. In addition, to improve the structure integrality, we propose a  position enhancement module to suppress incorrect nuclei boundary prediction. The experimental results demonstrate the effectiveness of our approaches and the superiority of SEINE. As a versatile method of structure modeling, SEINE provides a novel perspective for nuclei instance segmentation.

\bibliographystyle{IEEEtran}
\bibliography{tmi}

\end{document}